\documentclass[10pt,twocolumn,letterpaper]{article}

\usepackage[pagenumbers]{cvpr}

\usepackage{times}
\usepackage{epsfig}
\usepackage{graphicx}
\usepackage{amsmath}
\usepackage{amssymb}
\usepackage{xcolor}
\usepackage{booktabs}
\usepackage{cite}
\usepackage{microtype}
\usepackage{enumitem}
\usepackage[accsupp]{axessibility}

\usepackage{pifont}
\usepackage{ragged2e}
\usepackage{url}

\usepackage{textcomp}
\usepackage{stfloats}
\usepackage{verbatim}

\usepackage{enumerate}

\usepackage{xcolor}

\definecolor{brickred}{rgb}{0.8, 0.25, 0.33}
\definecolor{blueish}{rgb}{0.0, 0.3, 0.6}
\usepackage[pagebackref=false,breaklinks=true,letterpaper=true,colorlinks,bookmarks=false,citecolor=blueish,linkcolor=brickred]{hyperref}

\setitemize{noitemsep,topsep=0pt,parsep=0pt,partopsep=0pt}

\newcommand{\cmark}{\ding{51}}
\newcommand{\xmark}{\ding{55}}

\def\confName{CVPR}
\def\confYear{2024}

\title{GraphMLP: A Graph MLP-Like Architecture for 3D Human Pose Estimation}

\author{
Wenhao Li\textsuperscript{1} \quad
Mengyuan Liu\textsuperscript{1}\thanks{Corresponding Author.} \quad
Hong Liu\textsuperscript{1} \quad
Tianyu Guo\textsuperscript{1} \quad
Ti Wang\textsuperscript{1} \quad
Hao Tang\textsuperscript{1} \quad
Nicu Sebe\textsuperscript{2}
\\
\textsuperscript{1}{Peking University} \quad
\textsuperscript{2}{University of Trento} \\
{\tt\small \{wenhaoli,liumengyuan,hongliu,haotang\}@pku.edu.cn} \\
 {\tt\small \{levigty,tiwang\}@stu.pku.edu.cn} \quad  {\tt\small niculae.sebe@unitn.it}
}

\begin{document}
\maketitle

\begin{abstract}
Modern multi-layer perceptron (MLP) models have shown competitive results in learning visual representations without self-attention. However, existing MLP models are not good at capturing local details and lack prior knowledge of human body configurations, which limits their modeling power for skeletal representation learning. To address these issues, we propose a simple yet effective graph-reinforced MLP-Like architecture, named GraphMLP, that combines MLPs and graph convolutional networks (GCNs) in a \textit{global-local-graphical} unified architecture for 3D human pose estimation. GraphMLP incorporates the graph structure of human bodies into an MLP model to meet the domain-specific demand of the 3D human pose, while allowing for both local and global spatial interactions. Furthermore, we propose to flexibly and efficiently extend the GraphMLP to the video domain and show that complex temporal dynamics can be effectively modeled in a simple way with negligible computational cost gains in the sequence length. To the best of our knowledge, this is the first MLP-Like architecture for 3D human pose estimation in a single frame and a video sequence. Extensive experiments show that the proposed GraphMLP achieves state-of-the-art performance on two datasets, \emph{i.e.}, Human3.6M and MPI-INF-3DHP. Code and models are available at \url{https://github.com/Vegetebird/GraphMLP}.
\end{abstract}

\section{Introduction}
3D human pose estimation from images is important in numerous applications, such as action recognition, motion capture, and augmented/virtual reality. 
Most existing works solve this task by using a 2D-to-3D pose lifting method, which takes graph-structured 2D joint coordinates detected by a 2D keypoint detector as input~\cite{simplebaseline,wei2019view,kim2023mhcanonnet,hua2022weakly}. 
This is an inherently ambiguous problem since multiple valid 3D joint locations may correspond to the same 2D projection in the image space. 
However, it is practically solvable since 3D poses often lie on low-dimensional manifolds, which can provide important structural priors to mitigate the depth ambiguity~\cite{wang2014robust,ci2019optimizing}. 

\begin{figure}[tb]
\centering
\includegraphics[width=1.00\linewidth]{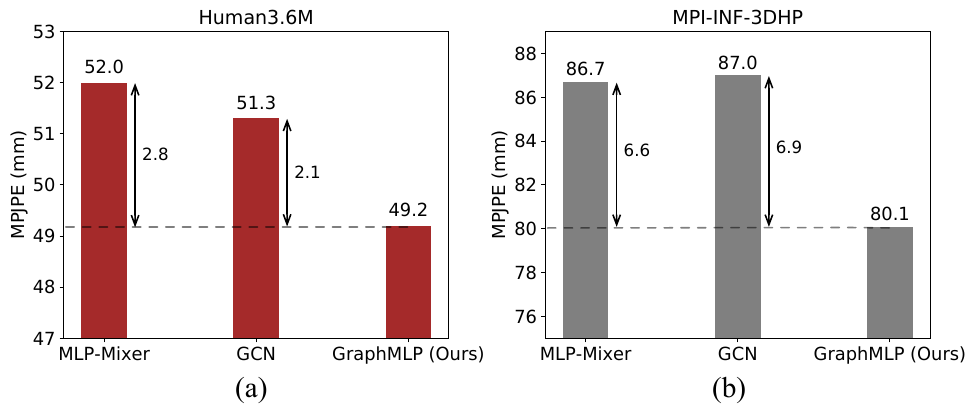}
\caption
{Performance comparison with MLP-Mixer~\cite{mlpmixer} and GCN~\cite{stgcn} on Human3.6M (a) and MPI-INF-3DHP (b) datasets. 
The proposed GraphMLP absorbs the advantages of modern MLPs and GCNs to effectively learn skeletal representations, consistently outperforming each of them.  
The evaluation metric is MPJPE (the lower the better). 
}
\label{fig:compare}
\end{figure}

\begin{figure*}[tb]
\centering
\includegraphics[width=0.9\linewidth]{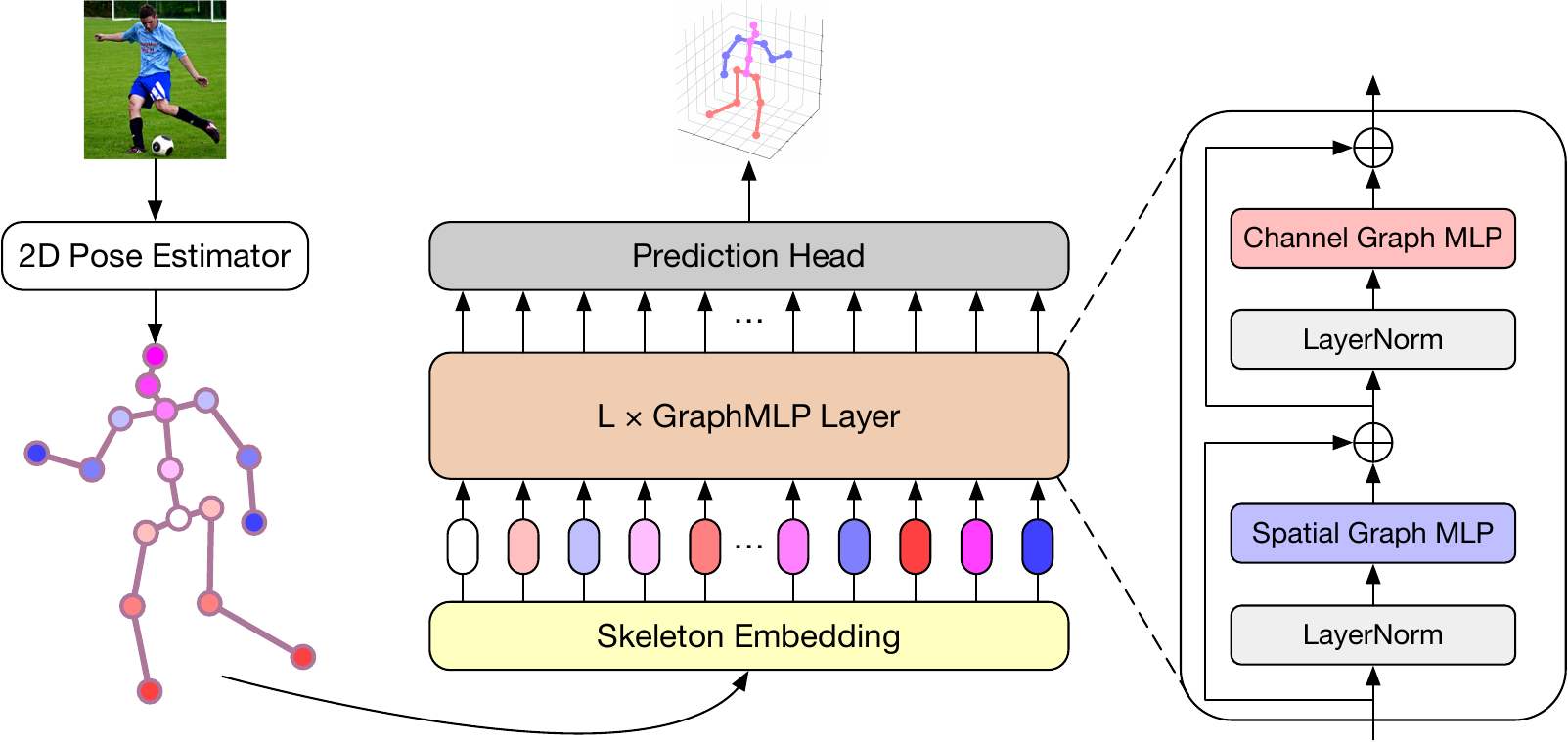}
\caption
{
Overview of the proposed GraphMLP architecture. 
The left illustrates the skeletal structure of the human body. 
The 2D joint inputs detected by a 2D pose estimator are sparse and graph-structured data.  
GraphMLP treats each 2D keypoint as an input token, linearly embeds each of them through the skeleton embedding, feeds the embedded tokens to GraphMLP layers, and finally performs regression on resulting features to predict the 3D pose via the prediction head. 
Each GraphMLP layer contains one spatial graph MLP (SG-MLP) and one channel graph MLP (CG-MLP). 
For easy illustration, we show the architecture using a single image as input. 
}
\label{fig:overview}
\end{figure*}

Early works attempted to employ fully connected networks (FCN)~\cite{simplebaseline} to lift the 2D joints into 3D space. 
However, the dense connection of FCN often results in overfitting and poor performance~\cite{zeng2021learning}. 
To address this problem, recent works consider that the skeleton of a human body can be naturally represented as the graph structure and utilize graph convolutional networks (GCNs) for this task~\cite{zhao2019semantic,ci2019optimizing,liu2020comprehensive}. 
Although GCN-based methods are effective at aggregating neighboring nodes to extract local features, they often suffer from limited receptive fields to obtain stronger representation power~\cite{zhao2022graformer}, as these methods usually rely on learning relationships between human body joints utilizing first-hop neighbors. 
However, global information between distant body joints is crucial for understanding the overall body posture and movement patterns~\cite{zhang2023learning,zhao2022graformer}. For instance, the position of the hand relative to the foot can indicate whether the person is standing, sitting, or lying down. 
One way to capture global information is by stacking multiple GCN layers, which expands the model's receptive fields to cover the entire kinematic chain of a human.
However, this leads to the over-smoothing issue where valuable information is lost in deeper layers~\cite{li2018deeper}. 

Recently, modern multi-layer perceptron (MLP) models (in particular MLP-Mixer~\cite{mlpmixer}) with global receptive fields provide new architectural designs in vision~\cite{lian2021mlp,liu2021pay,ren2021cascaded,chen2022cyclemlp}. 
The MLP model stacks only fully-connected layers without self-attention and consists of two types of blocks. 
The spatial MLP aggregates global information among tokens, and the channel MLP focuses on extracting features for each token. 
By stacking these two MLP blocks, it can be simply built with less inductive bias and achieves impressive performance in learning visual representations. 
This motivates us to explore an MLP-Like architecture for skeleton-based representation learning. 

However, there remain two critical challenges in adapting the MLP models from vision to skeleton:
\textbf{(i)} Despite their successes in vision, existing MLPs are less effective in modeling graph-structured data due to their simple connections among all nodes. 
Different from RGB images represented by highly dense pixels, skeleton inputs are inherently sparse and graph-structured data (see Fig.~\ref{fig:overview} (left)). 
Without incorporating the prior knowledge of human body configurations, the model is prone to learning spurious dependencies, which can lead to physically implausible poses~\cite{stgcn,zeng2021learning}. 
\textbf{(ii)} 
While such models are capable of capturing global interactions between distant body joints (such as the hand and foot) via their spatial MLPs, they may not be good at capturing local interactions due to the lack of careful designs for modeling relationships between adjacent joints (such as the head and neck). 
However, local information is also essential for 3D human pose estimation, as it can help the model to understand fine-grained movement details~\cite{xu2021graph,meshgraphormer}. 
For example, the movement of the head relative to the neck can indicate changes in gaze direction or subtle changes in posture. 

To overcome both limitations, we present GraphMLP, a new graph-reinforced MLP-Like architecture for 3D human pose estimation, as depicted in Fig.~\ref{fig:overview}. 
Our GraphMLP is conceptually simple yet effective: it builds a strong collaboration between modern MLPs and GCNs to construct a \textit{global-local-graphical} unified architecture for learning better skeletal representations. 
Specifically, GraphMLP mainly contains a stack of novel GraphMLP layers. 
Each layer consists of two Graph-MLP blocks where the spatial graph MLP (SG-MLP) and the channel graph MLP (CG-MLP) blocks are built by injecting GCNs into the spatial MLP and channel MLP, respectively. 
By combining MLPs and GCNs in a unified architecture, our GraphMLP is able to obtain the prior knowledge of human configurations encoded by the graph's connectivity and capture both local and global spatial interactions among body joints, hence yielding better performance.

Furthermore, we extend our GraphMLP from a single frame to a video sequence. 
Most existing video-based methods~\cite{videopose,poseformer,mhformer} typically model temporal information by treating each frame as a token or treating the time axis as a separated dimension. 
However, these methods often suffer from redundant computations that make little contribution to the final performance because nearby poses are similar~\cite{strided}.
Additionally, they are too computationally expensive to process long videos (\emph{e.g.}, 243 frames), thereby limiting their practical utility in real-world scenarios. 
To tackle these issues, we propose to utilize a simple and efficient video representation of pose sequences.  
This representation captures complex temporal dynamics by mixing temporal information in the feature channels and treating each joint as a token, offering negligible computational cost gains in the sequence length. 
It is also a unified and flexible representation that can accommodate arbitrary-length sequences (\emph{i.e.}, a single frame and variable-length videos). 

The proposed GraphMLP is evaluated on two challenging datasets, Human3.6M~\cite{ionescu2013human3} and MPI-INF-3DHP~\cite{mehta2017monocular}. 
Extensive experiments show the effectiveness and generalization ability of our approach, which advances the state-of-the-art performance for estimating 3D human poses from a single image. 
Its performance surpasses MGCN~\cite{zou2021modulated} by 1.4 $mm$ in mean per joint position error (MPJPE) on the Human3.6M dataset. 
Besides, as shown in Fig.~\ref{fig:compare}, it brings clear 6.6 $mm$ and 6.9 $mm$ improvements in MPJPE for the MLP model~\cite{mlpmixer} and the GCN model~\cite{stgcn} on the MPI-INF-3DHP dataset, respectively. 
Surprisingly, compared to video pose Transformer (\emph{e.g.}, PoseFormer~\cite{poseformer}), even with $5\times$ fewer computational costs, our MLP-Like architecture achieves better performance. 

Overall, our main contributions are summarized as follows:
\begin{itemize}
\item We present, to the best of our knowledge, the first MLP-Like architecture called GraphMLP for 3D human pose estimation. 
It combines the advantages of modern MLPs and GCNs, including globality, locality, and connectivity. 
\item The novel SG-MLP and CG-MLP blocks are proposed to encode the graph structure of human bodies within MLPs to obtain domain-specific knowledge about the human body while enabling the model to capture both local and global interactions. 
\item  A simple and efficient video representation is proposed to extend our GraphMLP to the video domain flexibly. 
This representation enables the model to effectively process arbitrary-length sequences with negligible computational cost gains. 
\item Extensive experiments demonstrate the effectiveness and generalization ability of the proposed GraphMLP and show new state-of-the-art results on two challenging datasets, \emph{i.e.}, Human3.6M~\cite{ionescu2013human3} and MPI-INF-3DHP~\cite{mehta2017monocular}. 
\end{itemize}

\section{Related Work} 
\noindent \textbf{3D Human Pose Estimation.}
There are mainly two categories to estimate 3D human poses. 
The first category of methods directly regresses 3D human joints from RGB images~\cite{li20143d,pavlakos2017coarse,qiu2023weakly,han2022single}. 
The second category is the 2D-to-3D pose lifting method~\cite{simplebaseline,videopose,hot}, which employs an off-the-shelf 2D pose detection as the front end and designs a 2D-to-3D lifting network using detected 2D poses as input. 
This lifting method can achieve state-of-the-art performance and has become the mainstream method due to its efficiency and effectiveness. 
For example, FCN~\cite{simplebaseline} shows that 3D poses can be regressed simply and effectively from 2D keypoints with fully-connected networks. 
TCN~\cite{videopose} extends the FCN to video by utilizing temporal convolutional networks to exploit temporal information from 2D pose sequences. 
Liu \emph{et al.}~\cite{liu2020attention} incorporate the attention mechanism into TCN to enhance the modeling of long-range temporal relationships across frames. 
SRNet~\cite{zeng2020srnet} proposes a split-and-recombine network that splits the human body joints into multiple local groups and recombines them with a low-dimensional global context. 
PoSynDA \cite{liu2023posynda} uses domain adaptation through multi-hypothesis pose synthesis for 3D human pose estimation. 

Since the physical skeleton topology can form a graph structure, recent progress has focused on employing graph convolutional networks (GCNs) to address the 2D-to-3D lifting problem. 
LCN~\cite{ci2019optimizing} introduces a locally connected network to improve
the representation capability of GCN. 
SemGCN~\cite{zhao2019semantic} allows the model to learn the semantic relationships among the human joints. 
MGCN~\cite{zou2021modulated} improves SemGCN by introducing a weight modulation and an affinity modulation. 

\noindent \textbf{Transformers in Vision.}
Recently, Transformer-based methods achieve excellent results on various computer vision tasks, such as image classification~\cite{vit,swin,wang2023aa}, object detection~\cite{dert,zhu2020deformable,zhao2022tracking}, and pose estimation~\cite{poseformer,chen2023hdformer,li2023multi}. 
The seminal work of ViT~\cite{vit} divides an image into $16 {\times} 16$ patches and uses a pure Transformer encoder to extract visual features. 
PoseFormer~\cite{poseformer} utilizes a pure Transformer-based architecture to model spatial and temporal relationships from videos. 
Strided Transformer~\cite{strided} incorporates strided convolutions into Transformers to aggregate information from local contexts for video-based 3D human pose estimation. 
HDFormer~\cite{chen2023hdformer} proposes a High-order Directed Transformer to utilize high-order information on a directed skeleton graph based on Transformer. 
RTPCA~\cite{li2023refined} introduces a temporal pyramidal compression-and-amplification design for enhancing temporal modeling in 3D human pose estimation. 
Mesh Graphormer~\cite{meshgraphormer} combines GCNs and attention layers in a serial order to capture local and global dependencies for human mesh reconstruction. 

Unlike~\cite{meshgraphormer}, we mainly investigate how to combine more efficient architectures (\emph{i.e.}, modern MLPs) and GCNs to construct a stronger architecture for 3D human pose estimation and adopt a parallel manner to make it possible to model local and global information at the same time. 
Moreover, different from previous video-based methods~\cite{poseformer,strided,mhformer} that treat each frame as a token for temporal modeling, we mix features across all frames and maintain each joint as a token, which makes the network to be economical and easy to train. 

\noindent \textbf{MLPs in Vision.}
Modern MLP models are proposed to reduce the inductive bias and computational cost by replacing the complex self-attentions of Transformers with spatial-wise linear layers~\cite{lian2021mlp,liu2021pay,shi2022polyp}. 
MLP-Mixer~\cite{mlpmixer} firstly proposes an MLP-Like model, which is a simple network architecture containing only pure MLP layers. 
Compared with FCN~\cite{simplebaseline} (\emph{i.e.}, conventional MLPs), this architecture introduces some modern designs, \emph{e.g.}, layer normalization (LN)~\cite{ba2016layer}, GELU~\cite{hendrycks2016gaussian}, mixing spatial information. 
Moreover, ResMLP~\cite{touvron2021resmlp} proposes a purely MLP architecture with the Affine transformation. 
CycleMLP~\cite{chen2022cyclemlp} proposes a cycle fully-connected layer to aggregate spatial context information and deal with variable
input image scales. 
However, these modern MLP models have not yet been applied to 3D human pose estimation. 

Inspired by their successes in vision, we first attempt to explore how MLP-Like architectures can be used for 3D human pose estimation in non-Euclidean skeleton data. 
The difference between our approach and the existing MLPs is that we introduce the inductive bias of the physical skeleton topology by combining MLP models with GCNs, providing more physically plausible and accurate estimations. 
We further investigate applying MLP-Like architecture in video and design an efficient video representation, which is seldom studied. 

\section{Proposed GraphMLP}
Fig.~\ref{fig:overview} illustrates the overall architecture of the proposed GraphMLP. 
Our approach takes 2D joint locations $P \in \mathbb{R}^{N \times 2}$ estimated by an off-the-shelf 2D pose detector as input and outputs predicted 3D poses $\widetilde{X} \in \mathbb{R}^{N \times 3}$, where $N$ is the number of joints. 
The proposed GraphMLP architecture consists of a skeleton embedding module, a stack of $L$ identical GraphMLP layers, and a prediction head module. 
The core operation of GraphMLP architecture is the GraphMLP layer, each of which has two parts: a spatial graph MLP (SG-MLP) block and a channel graph MLP (CG-MLP) block. 
Our GraphMLP has a similar architecture to the original MLP-Mixer~\cite{mlpmixer}, but we incorporate graph convolutional networks (GCNs) into the model to meet the domain-specific requirement of the 3D human pose estimation and learn the local and global interactions of human body joints. 

\begin{figure}[t]
\centering
\includegraphics[width=1.00\linewidth]{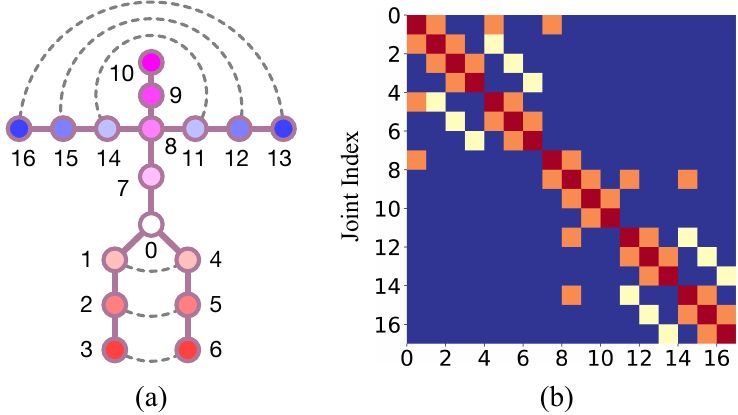}
\caption
{
\textbf{(a)} The human skeleton graph in physical and symmetrical connections. 
\textbf{(b)} The adjacency matrix used in the GCN blocks of GraphMLP. 
Different colors denote the different types of bone connections. 
}
\label{fig:graph}
\end{figure}

\subsection{Preliminary}
\label{sec:preliminary}
In this subsection, we briefly introduce the preliminaries in GCNs and modern MLPs. 

\subsubsection{Graph Convolutional Networks}
Let $G=(V, A)$ denotes a graph where $V$ is a set of $N$ nodes and $A \in \{0, 1\}^{N \times N}$ is the adjacency matrix encoding the edges between the nodes. 
Given a $(\ell-1)^{th}$ layer feature $X_{\ell-1} \in \mathbb{R}^{N \times C}$ with $C$ dimensions, a generic GCN~\cite{kipf2016semi} layer, which is used to aggregate features from neighboring nodes, can be formulated as:
\begin{equation}
  X_{\ell}=\widetilde{D}^{-\frac{1}{2}}  \widetilde{A} \widetilde{D}^{-\frac{1}{2}} X_{\ell-1} W,
\end{equation}
where $W \in \mathbb{R}^{C \times C^{\prime}}$ is the learnable weight matrix and $\widetilde{A} = A+I$ is the adjacency matrix with added self-connections. 
$I$ is the identity matrix, $\widetilde{D}$ is a diagonal matrix, and $\widetilde{D}^{i i}=\sum_{j} \tilde{A}^{i j}$. 

The design principle of the GCN block is similar to~\cite{stgcn}, but we adopt a simplified single-frame version that contains only one graph layer. 
For the GCN blocks of our GraphMLP, the nodes in the graph denote the joint locations of the human body in Fig.~\ref{fig:graph} (a), and the adjacency matrix represents the bone connections between every two joints for node information passing in Fig.~\ref{fig:graph} (b), \emph{e.g.}, the rectangle of 1st row and 2nd column denotes the connection between joint 0 and joint 1. 
In addition, the different types of bone connections use different kernel weights following~\cite{stgcn}. 

\subsubsection{Modern MLPs}
MLP-Mixer is the first modern MLP model proposed in~\cite{mlpmixer}. 
It is a simple and attention-free architecture that mainly consists of a spatial MLP and a channel MLP, as illustrated in Fig.~\ref{fig:layer} (a). 
The spatial MLP aims to transpose the spatial axis and channel axis of tokens to mix spatial information. Then the channel MLP processes tokens in the channel dimension to mix channel information. 
Let $X_{\ell-1} \in \mathbb{R}^{N \times C}$ be an input feature, an MLP-Mixer layer can be calculated as:
\begin{align}
  X^{\prime}_{\ell} &= X_{\ell-1} + \operatorname{Spatial-MLP}(\operatorname{LN}(X_{\ell-1}^{\top}))^{\top}, 
  \label{equ:mlpmixer_spatial} \\
  X_{\ell} &= X^{\prime}_{\ell} + \operatorname{Channel-MLP}(\operatorname{LN}(X^{\prime}_{\ell})),
  \label{equ:mlpmixer_channel}
\end{align}
where $\operatorname{LN}(\cdot)$ is layer normalization (LN)~\cite{ba2016layer} and $\top$ is the matrix transposition. 
Both spatial and channel MLPs contain two linear layers and a GELU~\cite{hendrycks2016gaussian} non-linearity in between. 

We adopt this MLP model as our baseline, which is similar to~\cite{mlpmixer}, but we transpose the tokens before LN in the spatial MLP (\emph{i.e.}, normalize tokens along the spatial dimension). 
However, such a simple MLP model neglects to extract fine-grained local details and lacks prior knowledge about the human configurations, which are perhaps the bottlenecks restricting the representation ability of MLP-Like architectures for learning skeletal representations. 

\subsection{Network Architecture}
In this work, we present a novel GraphMLP built upon the MLP-Mixer described in~\cite{mlpmixer} to overcome the aforementioned limitations of existing MLP models. 
Below we elaborate on each module used in GraphMLP and provide its detailed implementations. 

\subsubsection{Skeleton Embedding}
Raw input data are mapped to latent space via the skeleton embedding module. 
Given the input 2D pose $P \in \mathbb{R}^{N \times 2}$ with $N$ body joints, we treat each joint as an input token. 
These tokens are projected to the high-dimension token feature $X_{0} \in \mathbb{R}^{N \times C}$ by a linear layer, where $C$ is the hidden size. 

\begin{figure}[tb]
  \centering
  \includegraphics[width=1.00\linewidth]{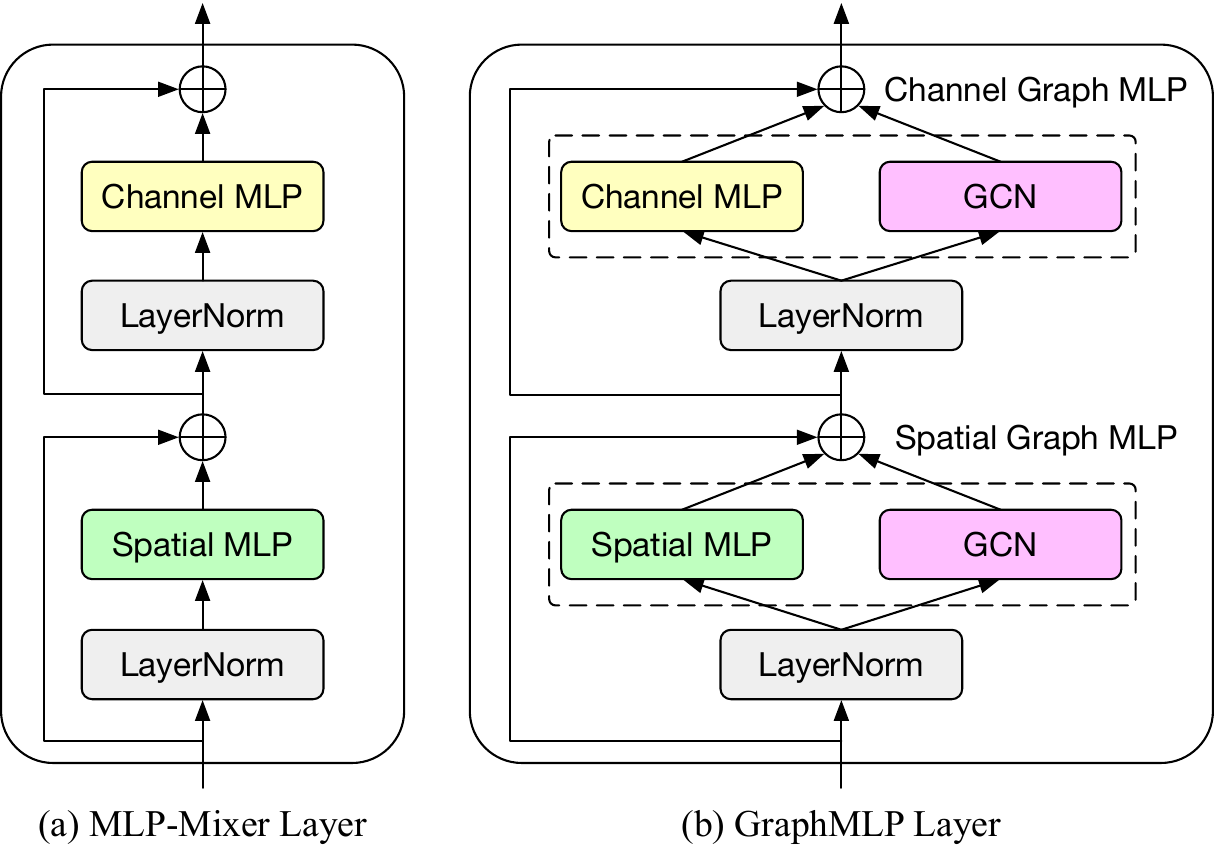}
  \caption
  {
    Comparison of MLP Layers. 
    \textbf{(a)} MLP-Mixer Layer~\cite{mlpmixer}. 
    \textbf{(b)} Our GraphMLP Layer.
    Compared with MLP-Mixer, our GraphMLP incorporates graph structural priors into the MLP model via GCN blocks. 
    The MLPs and GCNs are in a paralleled design to model both local and global interactions. 
  }
  \label{fig:layer}
\end{figure}

\subsubsection{GraphMLP Layer}
The existing MLP simply connects all nodes but does not take advantage of graph structures, making it less effective in handling graph-structured data. 
To tackle this issue, we introduce the GraphMLP layer that unifies globality, locality, and connectivity in a single layer. 
Compared with the original MLP-Mixer in Fig.~\ref{fig:layer} (a), the main difference of our GraphMLP layer (Fig.~\ref{fig:layer} (b)) is that we utilize GCNs for local feature communication. 
This modification retains the domain-specific knowledge of human body configurations, which induces an inductive bias enabling the GraphMLP to perform very well in skeletal representation learning. 

Specifically, our GraphMLP layer is composed of an SG-MLP and a CG-MLP. 
The SG-MLP and CG-MLP are built by injecting GCNs into the spatial MLP and channel MLP (mentioned in Sec.~\ref{sec:preliminary}), respectively. 
To be more specific, the SG-MLP contains a spatial MLP block and a GCN block. 
These blocks process token features in parallel, where the spatial MLP extracts features among tokens with a global receptive field, and the GCN block focuses on aggregating local information between neighboring joints. 
The CG-MLP has a similar architecture to SG-MLP but replaces the spatial MLP with the channel MLP and has no matrix transposition. 
Based on the above description, the MLP layers in Eq.~(\ref{equ:mlpmixer_spatial}) and Eq.~(\ref{equ:mlpmixer_channel}) are modified to process tokens as:
\begin{equation}
  \begin{aligned}
    X^{\prime}_{\ell} = X_{\ell-1} &+ \operatorname{Spatial-MLP}(\operatorname{LN}(X_{\ell-1}^{\top}))^{\top} \\
    &+ \operatorname{GCN}(\operatorname{LN}({X_{\ell-1}^{\top}})^{\top}), 
  \end{aligned}
\end{equation}
\begin{equation}
  \begin{aligned}
    X_{\ell} = X^{\prime}_{\ell} &+ \operatorname{Channel-MLP}(\operatorname{LN}(X^{\prime}_{\ell})) \\
    & + \operatorname{GCN}(\operatorname{LN}({X^{\prime}_{\ell}})),
  \end{aligned}
\end{equation}
where $\operatorname{GCN}(\cdot)$ denotes the GCN block, $\ell \in [1, \ldots, L]$ is the index of GraphMLP layers. 
Here $X^{\prime}_{\ell}$ and $X_{\ell}$ are the output features of the SG-MLP and the CG-MLP for block $\ell$, respectively.  

\subsubsection{Prediction Head}
Different from~\cite{vit,mlpmixer} that use a classifier head to do classification, our prediction head performs regression with a linear layer. 
It is applied on the extracted features $X_{L} \in \mathbb{R}^{N \times C}$ of the last GraphMLP layer to predict the final 3D pose $\widetilde{X} \in \mathbb{R}^{N \times 3}$. 

\subsubsection{Loss Function}
To train our GraphMLP, we apply an $L_{2}$-norm loss to calculate the difference between prediction and ground truth. 
The model is trained in an end-to-end fashion, and the $L_{2}$-norm loss is defined as follows:
\begin{equation}
  \mathcal{L}=\sum_{n=1}^{N} \left\|J_{n}-\widetilde{X}_{n}\right\|_{2},
\end{equation}
where $\widetilde{X}_{n}$ and $J_{n}$ are the predicted and ground truth 3D coordinates of joint $n$, respectively. 

\begin{figure}[tb]
\centering
\includegraphics[width=0.65\linewidth]{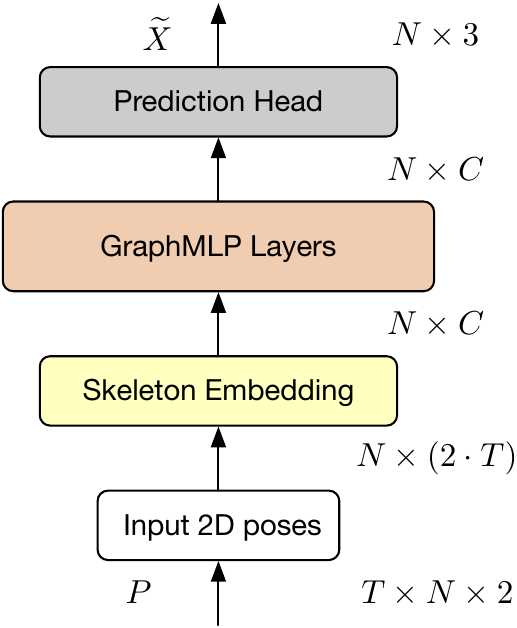}
\caption
{
    Illustration of the process of GraphMLP in the video domain. 
}
\label{fig:video}
\end{figure}

\subsection{Extension in the Video Domain}
To extend our GraphMLP for capturing temporal information, we introduce a simple video representation that changes the skeleton embedding module in the original architecture.  
Specifically, given a 2D pose sequence $P \in \mathbb{R}^{T \times N \times 2}$ with $T$ frames and $N$ joints, we first concatenate features between the coordinates of $x$-$y$ axis and all frames for each joint into $P^{\prime} \in \mathbb{R}^{N \times (2 \cdot T)}$, and then fed it into a linear layer $W \in \mathbb{R}^{(2 \cdot T) \times C}$ to map the tokens and get $X_{0} \in \mathbb{R}^{N \times C}$. 
Subsequently, each joint is treated as an input token and fed into the GraphMLP layers and prediction head module to output the 3D pose of the center frame $\widetilde{X} \in \mathbb{R}^{N \times 3}$. 
These processes of GraphMLP in the video domain are illustrated in Figure~\ref{fig:video}. 

This is also a unified and flexible representation strategy that can process arbitrary-length sequences, \emph{e.g.}, a single frame with $T=1$, and variable-length videos with $T=27,81,243$. 
Interestingly, we find that using a single linear layer to encode temporal information can achieve competitive performance without explicitly temporal modeling by treating each frame as a token. 
More importantly, this representation is efficient since the amount of $N$ is small (\emph{e.g.}, 17 joints). 
The increased computational costs from single frame to video sequence are only in one linear layer that weights are $W \in \mathbb{R}^{(2 \cdot T) \times C}$ with linear computational complexity to sequence length, which can be neglected. 
Meanwhile, the computational overhead and the number of parameters in the GraphMLP layers and the prediction head module are the same for different input sequences. 

\section{Experiments}
In this section, we first introduce experimental settings and implementation details for evaluation. 
Then, we compare the proposed GraphMLP with state-of-the-art methods. 
We also conduct detailed ablation studies on the importance of designs in our proposed approach. 

\subsection{Datasets and Evaluation Metrics}
\noindent \textbf{Human3.6M.} Human3.6M~\cite{ionescu2013human3} is the largest benchmark for 3D human pose estimation. 
It contains 3.6 million video frames captured by a motion capture system in an indoor environment, where 11 professional actors perform 15 actions such as greeting, phoning, and sitting. 
Following previous works~\cite{simplebaseline,xu2021graph,zeng2021learning}, our model is trained on five subjects (S1, S5, S6, S7, S8) and tested on two subjects (S9, S11). 
We report our performance using two evaluation metrics. 
One is the mean per joint position error (MPJPE), referred to as Protocol~\#1, which calculates the mean Euclidean distance in millimeters between the predicted and the ground truth joint coordinates. 
The other is the PA-MPJPE that measures the MPJPE after Procrustes analysis (PA)~\cite{gower1975generalized} and is referred to as Protocol~\#2. 

\noindent \textbf{MPI-INF-3DHP.}
MPI-INF-3DHP~\cite{mehta2017monocular} is a large-scale 3D human pose dataset containing both indoor and outdoor scenes. 
Its test set consists of three different scenes: studio with green screen (GS), studio without green screen (noGS), and outdoor scene (Outdoor).
Following previous works~\cite{ci2019optimizing,zeng2021learning,zou2021modulated}, we use the Percentage of Correct Keypoint (PCK) with the threshold of 150 $mm$ and Area Under Curve (AUC) for a range of PCK thresholds as evaluation metrics. 
To verify the generalization ability of our approach, we directly apply the model trained on Human3.6M to the test set of this dataset. 

\subsection{Implementation Details}
In our implementation, the proposed GraphMLP is stacked by $L = 3$ GraphMLP layers with hidden size $C = 512$, the MLP dimension of CG-MLP, \textit{i.e.}, $D_{C} = 1024$, and the MLP dimension of SG-MLP, \textit{i.e.}, $D_{S} = 256$. 
The whole framework is trained in an end-to-end fashion from scratch on a single NVIDIA RTX 2080 Ti GPU. 
The learning rate starts from 0.001 with a decay factor of 0.95 utilized in each epoch and 0.5 utilized per 5 epochs. 
We follow the generic data augmentation (horizontal flip augmentation) in~\cite{videopose,stgcn,zou2021modulated}. 
Following~\cite{xu2021graph,zou2021modulated,zeng2021learning}, we use 2D joints detected by cascaded pyramid network (CPN)~\cite{chen2018cascaded} for Human3.6M and 2D joints provided by the dataset for MPI-INF-3DHP. 

\begin{table*}[tb]
\centering
\caption
{
Quantitative comparison with state-of-the-art single-frame methods on Human3.6M under Protocol \#1 and Protocol \#2. 
Detected 2D keypoints are used as input.  
$\S$ - adopts the same refinement module as~\cite{stgcn,zou2021modulated,hassan2023regular}. 
}
\resizebox{\textwidth}{!}{
\begin{tabular}{@{}l|ccccccccccccccc|c@{}}
\toprule
\textbf{Protocol \#1} & Dir. & Disc & Eat & Greet & Phone & Photo & Pose & Purch. & Sit & SitD. & Smoke & Wait & WalkD. & Walk & WalkT. & Avg.\\
\midrule

FCN~\cite{simplebaseline} ICCV'17 &51.8 &56.2 &58.1 &59.0 &69.5 &78.4 &55.2 &58.1 &74.0 &94.6 &62.3 &59.1 &65.1 &49.5 &52.4 &62.9 \\

TCN~\cite{videopose} CVPR'19 &47.1& 50.6& 49.0& 51.8 &53.6 &61.4& 49.4 &47.4 &59.3 &67.4 &52.4& 49.5& 55.3& 39.5& 42.7& 51.8 \\

ST-GCN~\cite{stgcn} ICCV'19$\S$&46.5 &48.8 &47.6& 50.9& 52.9 &61.3 &48.3 &45.8 &59.2 &64.4& 51.2& 48.4& 53.5& 39.2& 41.2& 50.6 \\

SRNet~\cite{zeng2020srnet} ECCV'20 &44.5& {48.2} &47.1 &{47.8} &51.2 &{56.8} &50.1& {45.6}& 59.9 &66.4 &52.1 &{45.3} &54.2 &39.1 &40.3 &49.9 \\

GraphSH~\cite{xu2021graph} CVPR'21 &45.2 &49.9 &47.5 &50.9 &54.9 &66.1 &48.5 &46.3 &59.7 &71.5 &51.4 &48.6 &53.9 &39.9 &44.1 &51.9 \\

MGCN~\cite{zou2021modulated} ICCV'21$\S$ &45.4 &49.2 &45.7 &49.4 &{50.4} &58.2 &47.9 &46.0 &57.5 &63.0 &49.7 &46.6 &52.2 &38.9 &40.8 &49.4 \\

GraFormer~\cite{zhao2022graformer} CVPR'22 &45.2 &50.8 &48.0 &50.0 &54.9 &65.0 &48.2 &47.1 &60.2 &70.0 &51.6 &48.7 &54.1 &39.7 &43.1 &51.8 \\

UGRN \cite{li2023pose} AAAI'23 & 47.9 & 50.0 & 47.1 & 51.3 & 51.2 & 59.5 & 48.7 & 46.9 &{56.0} & 61.9 & 51.1 & 48.9 & 54.3 & 40.0 & 42.9 & 50.5 \\

RS-Net \cite{hassan2023regular} TIP'23$\S$  &44.7 &48.4 &44.8 &49.7 &49.6 &58.2 &47.4 &44.8 &55.2 &59.7 &49.3 &46.4 &51.4 &38.6 &40.6 &48.6 \\

\midrule

GraphMLP (Ours) &{45.4} &{50.2} &{45.8} &{49.2} &{51.6} &{57.9} &{47.3} &{44.9} &{56.9} &{61.0} &{49.5} &{46.9} &{53.2} &{37.8} &{39.9} &{49.2} \\

GraphMLP (Ours)$\S$ &{43.7} &{49.3} &{45.5} &{47.9} &{50.5} &{56.0} &{46.3} &{44.1} &{55.9} &{59.0} &{48.4} &{45.7} &{51.2} &{37.1} &{39.1} &\textbf{48.0} \\

\toprule
\textbf{Protocol \#2} & Dir. & Disc & Eat & Greet & Phone & Photo & Pose & Purch. & Sit & SitD. & Smoke & Wait & WalkD. & Walk & WalkT. & Avg. \\
\midrule

FCN~\cite{simplebaseline} ICCV'17 &39.5 &43.2 &46.4 &47.0 &51.0 &56.0 &41.4 &40.6 &56.5 &69.4 &49.2 &45.0 &49.5 &38.0 &43.1 &47.7 \\

TCN~\cite{videopose} CVPR'19 &36.0 &38.7 &38.0 &41.7 &40.1 &45.9 &37.1 &35.4 &46.8 &53.4 &41.4 &36.9 &43.1 &30.3 &34.8 &40.0 \\

ST-GCN~\cite{stgcn} ICCV'19$\S$ &36.8 &38.7 &38.2 &41.7 &40.7 &46.8 &37.9 &35.6 &47.6 &51.7 &41.3 &36.8 &42.7 &31.0 &34.7 &40.2 \\

SRNet~\cite{zeng2020srnet} ECCV'20 &35.8 &39.2 &{36.6} &{36.9} &39.8 &45.1 &38.4 &36.9 &47.7 &54.4 &{38.6} &36.3 &{39.4} &30.3 &35.4 &39.4 \\

MGCN~\cite{zou2021modulated} ICCV'21$\S$ &35.7 &38.6 &{36.3} &40.5 &{39.2} &44.5 &37.0 &35.4 &46.4 &51.2 &40.5 &35.6 &41.7 &30.7 &33.9 &39.1 \\

SGNN~\cite{zeng2021learning} ICCV'21 &{33.9} &{37.2} &{36.8} &{38.1} &{38.7} &{43.5} &{37.8} &{35.0} &{47.2} &{53.8} &{40.7} &{38.3} &{41.8} &{30.1} &{31.4} &{39.0} \\

RS-Net \cite{hassan2023regular} TIP'23$\S$  &35.5 &38.3 &36.1 &40.5 &39.2 &44.8 &37.1 &34.9 &45.0 &49.1 &40.2 &35.4 &41.5 &31.0 &34.3 &38.9  \\

\midrule
GraphMLP (Ours) &{35.0} &{38.4} &{36.6} &{39.7} &{40.1} &{43.9} &{35.9} &{34.1} &{45.9} &{48.6} &{40.0} &{35.3} &{41.6} &{30.0} &{33.3} &{38.6} \\

GraphMLP (Ours)$\S$ &{35.1} &{38.2} &{36.5} &{39.8} &{39.8} &{43.5} &{35.7} &{34.0} &{45.6} &{47.6} &{39.8} &{35.1} &{41.1} &{30.0} &{33.4} &\textbf{38.4} \\

\toprule
\end{tabular}
}
\label{table:h36m}
\end{table*}

\subsection{Comparison with State-of-the-Art Methods}
\noindent \textbf{Comparison with Single-frame Methods.}
Table~\ref{table:h36m} reports the performance comparison between our GraphMLP and previous state-of-the-art methods that take a single frame as input on Human3.6M. 
It can be seen that our approach reaches 49.2 $mm$ in MPJPE and 38.6 $mm$ in PA-MPJPE, which outperforms previous methods. 
Note that some works~\cite{stgcn,zou2021modulated,hassan2023regular} adopt a pose refinement module to boost the performance further. 
Compared with them, GraphMLP achieves lower MPJPE (48.0 $mm$), surpassing MGCN~\cite{zou2021modulated} by a large margin of 1.4 $mm$ error reduction (relative 3\% improvements). 

\begin{table*}[t]
\centering
\caption
{
Quantitative comparison with state-of-the-art single-frame methods on Human3.6M under Protocol \#1. 
Ground truth 2D keypoints are used as input. 
} 
\resizebox{\textwidth}{!}{
\begin{tabular}{@{}l|ccccccccccccccc|c@{}}
\toprule
\textbf{Protocol \#1} & Dir. & Disc & Eat & Greet & Phone & Photo & Pose & Purch. & Sit & SitD. & Smoke & Wait & WalkD. & Walk & WalkT. & Avg.\\
\midrule

FCN~\cite{simplebaseline} ICCV'17 &37.7 &44.4 &40.3 &42.1& 48.2& 54.9 &44.4 &42.1 &54.6& 58.0 &45.1 &46.4 &47.6 &36.4 &40.4& 45.5\\

SemGCN~\cite{zhao2019semantic} CVPR'19 &37.8 &49.4 &37.6 &40.9 &45.1 &41.4 &40.1& 48.3& 50.1 &42.2& 53.5 &44.3 &40.5 &47.3& 39.0 &43.8 \\

ST-GCN~\cite{stgcn} ICCV'19 &{33.4} &39.0 &33.8 &37.0 &38.1 &47.3 &39.5 &37.3 &43.2 &46.2 &37.7 &38.0 &38.6 &30.4 &32.1 &38.1\\

SRNet~\cite{zeng2020srnet} ECCV'20 &35.9 &{36.7} &{29.3} &34.5 &36.0 &42.8 &37.7 &31.7 &40.1 &44.3 &35.8 &37.2 &36.2 &33.7 &34.0 &36.4  \\

GraphSH~\cite{xu2021graph} CVPR'21 &35.8 &38.1 &31.0 &35.3 &35.8 &43.2 &{37.3} &{31.7} &38.4 &45.5 &35.4 &{36.7} & 36.8 &{27.9} &30.7 &35.8 \\

GraFormer~\cite{zhao2022graformer} CVPR'22 &{32.0} &38.0 &30.4 &34.4 &34.7 &43.3 &{35.2} &{31.4} &38.0 &46.2 &34.2 &{35.7} &36.1 &{27.4} &30.6 &35.2 \\

PHGANet \cite{zhang2023learning} IJCV'23 &32.4 &36.5 &30.1 &33.3 &36.3 &43.5 &36.1 &30.5 &37.5 &45.3 &33.8 &35.1 &35.3 &27.5 &30.2 &34.9 \\

\midrule

GraphMLP (Ours) &{32.2} &{38.2} &{29.3} &{33.4} &{33.5} &{38.1} &{38.2} &{31.7} &{37.3} &{38.5} &{34.2} &{36.1} &{35.5} &{28.0} &{29.3} &\textbf{34.2} \\

\toprule
\end{tabular}
}
\label{table:gt}
\end{table*}

\begin{table}[tb]
\centering
\scriptsize
\caption
{
Performance comparison with state-of-the-art single-frame methods on MPI-INF-3DHP. 
}
\setlength{\tabcolsep}{1.15mm} 
\begin{tabular}{@{}l|cccccc@{}}
  \toprule [1pt]
Method &GS $\uparrow$ &noGS $\uparrow$ &Outdoor $\uparrow$ &All PCK $\uparrow$ &All AUC $\uparrow$ \\

\midrule [0.5pt]
FCN~\cite{simplebaseline} ICCV'17 &49.8 &42.5 &31.2 &42.5 &17.0 \\
LCN~\cite{ci2019optimizing} ICCV'19 &74.8 &70.8 &77.3 &74.0 &36.7 \\
SGNN~\cite{zeng2021learning} ICCV'21 &- &- &84.6 &82.1 &46.2 \\
MGCN~\cite{zou2021modulated} ICCV'21 &86.4 &86.0 &85.7 &86.1 &53.7 \\
GraFormer~\cite{zhao2022graformer} CVPR'22 &80.1 &77.9 &74.1 &79.0 &43.8 \\
UGRN~\cite{li2023pose} AAAI'23 & 86.2 & 84.7 & 81.9 & 84.1 & 53.7 \\
RS-Net \cite{hassan2023regular} TIP'23 &- &- &- &85.6 &53.2  \\
\midrule [0.5pt]
GraphMLP (Ours) &\textbf{87.3} &\textbf{87.1} &\textbf{86.3} &\textbf{87.0} &\textbf{54.3} \\

\toprule [1pt]
\end{tabular}
\label{table:3dhp}
\end{table}

Due to the uncertainty of 2D detections, we also report results using ground truth 2D keypoints as input to explore the upper bound of the proposed approach. 
As shown in Table~\ref{table:gt}, our GraphMLP obtains substantially better performance when given precise 2D joint information and attains state-of-the-art performance, which indicates its effectiveness. 

Table~\ref{table:3dhp} further compares our GraphMLP against previous state-of-the-art single-frame methods on cross-dataset scenarios. 
We only train our model on the Human3.6M dataset and test it on the MPI-INF-3DHP dataset. 
The results show that our approach obtains the best results in all scenes and all metrics, consistently surpassing other methods. 
This verifies the strong generalization ability of our approach to unseen scenarios. 

\begin{table}[t]
\centering
\scriptsize
\caption
{ 
Quantitative comparisons with video-based methods on Human3.6M under MPJPE.
CPN and GT denote the inputs of 2D poses detected by CPN and ground truth 2D poses, respectively. 
}
\setlength{\tabcolsep}{7.70mm} 
\begin{tabular}{l|cc}
\toprule [1pt]
Method &CPN &GT \\
\midrule [0.5pt]
ST-GCN~\cite{stgcn} ($T {=} 9$) &48.8 &37.2 \\
TCN~\cite{videopose} ($T{=}243$) &46.8 &37.8 \\
SRNet~\cite{zeng2020srnet} ($T{=}243$) &44.8 &32.0 \\
PoseFormer~\cite{poseformer} ($T{=}81$) &44.3 &31.3 \\
Anatomy3D~\cite{chen2021anatomy} 
($T{=}243$) &44.1 &32.3 \\
\midrule [0.5pt]
Baseline ($T{=}243$) &48.4 &35.2 \\

GraphMLP (Ours, $T{=}243$) &\textbf{43.8} &\textbf{30.3} \\

\toprule [1pt]
\end{tabular}
\label{table:video}
\end{table}

\begin{table}[t]
\centering
\scriptsize
\caption
{ 
  Comparison of parameters, FLOPs, and MPJPE with PoseFormer~\cite{poseformer}, MixSTE~\cite{mixste}, STCFormer~\cite{stcformer}, and MotionAGFormer~\cite{motionagformer} in different input frames on Human3.6M. 
  Frame per second (FPS) was computed on a single GeForce RTX 3090 GPU. 
}
\setlength{\tabcolsep}{1.60mm} 
\begin{tabular}{l|ccccc}
\toprule [1pt]
Model &$T$ &Param (M) &FLOPs (M) &MPJPE (mm) $\downarrow$ \\
\midrule [0.5pt]  
PoseFormer~\cite{poseformer} &1 &9.56 &{20} &51.3 \\
PoseFormer~\cite{poseformer} &27 &9.57 &541 &47.0 \\
PoseFormer~\cite{poseformer} &81 &9.60 &1625 &44.3 \\
PoseFormer~\cite{poseformer} &243 &9.69 &4874 &- \\

\midrule [0.5pt]

MixSTE~\cite{mixste} &1 &33.66 &1141 &51.1 \\
MixSTE~\cite{mixste} &27 &33.67 &30805 &45.1 \\
MixSTE~\cite{mixste} &81 &33.70 &92416 &42.4 \\
MixSTE~\cite{mixste} &243 &33.78 &277248 &40.9 \\

\midrule [0.5pt]

STCFormer~\cite{stcformer} &1 &- &- &- \\
STCFormer~\cite{stcformer} &27 &4.75 &4347 &44.1 \\
STCFormer~\cite{stcformer} &81 &4.75 &13041 &42.0 \\
STCFormer~\cite{stcformer} &243 &18.93 &156215 &40.5 \\

\midrule [0.5pt]

MotionAGFormer~\cite{motionagformer} &1 &- &- &- \\
MotionAGFormer~\cite{motionagformer} &27 &2.24 &2023 &45.1 \\
MotionAGFormer~\cite{motionagformer} &81 &4.82 &13036 &42.5 \\
MotionAGFormer~\cite{motionagformer} &243 &19.01 &155634 &38.4 \\

\midrule [0.5pt]

GraphMLP (Ours) &1 &{9.49} &348 &49.2 & \\
GraphMLP (Ours) &27 &{9.51} &{349} &45.5 & \\
GraphMLP (Ours) &81 &{9.57} &{351} &44.5 & \\
GraphMLP (Ours) &243 &{9.73} &{356} &43.8 & \\

\toprule [1pt]
\end{tabular}
\label{table:poseformer}
\end{table}

\noindent \textbf{Comparison with Video-based Methods.}
As shown in Table~\ref{table:video}, our method achieves outstanding performance against video-based methods in both CPN and GT inputs. 
The proposed method surpasses our baseline model (\emph{i.e.}, MLP-Mixer~\cite{mlpmixer}) by a large margin of 4.6 $mm$ (10\% improvements) with CPN inputs and 4.9 $mm$ (14\% improvements) with GT inputs.  
These results further demonstrate the effectiveness of our GraphMLP, which combines MLPs and GCNs to learn better skeleton representations. 
Compared with the most related work, Poseformer~\cite{poseformer}, a self-attention-based architecture, our GraphMLP, such a self-attention free architecture, improves the results from 31.3 $mm$ to 30.3 $mm$ with GT inputs (3\% improvements). 
Besides, our simple video representation, which uses a single linear layer to encode temporal information instead of being designed explicitly for temporal enhancement like previous works~\cite{videopose,chen2021anatomy,poseformer}, is also capable of performing well. 
This indicates that our method can alleviate the issue of video redundancy by compressing the video information into a single vector, leading to impressive results. 

Table~\ref{table:poseformer} further reports the comparison of parameters, FLOPs, and MPJPE with PoseFormer~\cite{poseformer}, MixSTE~\cite{mixste}, STCFormer~\cite{stcformer}, and MotionAGFormer~\cite{motionagformer} in different input frames. 
Surprisingly, compared with PoseFormer, our method requires only 22\% FLOPs (356M vs. 1625M) while achieving better performance. 
Although MixSTE, STCFormer, and MotionAGFormer achieve better performance with the 243-frame model, they respectively require $779 \times$ (356M vs. 277248M), $ 439\times$ (356M vs. 156215M), and $437 \times$ (356M vs. 155634M) more FLOPs than ours, which leads to a time-consuming training process and difficulties in deployment. 
Note that our method offers negligible FLOPs gains in the sequence length, which allows it easy to deploy in real-time applications and has great potential for better results with longer sequences. 
These results demonstrate that our GraphMLP in video reaches competitive performance with fewer computational costs and can serve as a strong baseline for video-based 3D human pose estimation. 

\subsection{Ablation Study}
\label{sec:ablation}

The large-scale ablation studies with 2D detected inputs on the Human3.6M dataset are conducted to investigate the effectiveness of our model (using the single-frame model). 

\noindent \textbf{Model Configurations.}
We start our ablation studies by exploring the GraphMLP on different hyper-parameters. 
The results are shown in Table~\ref{table:parameters}. 
It can be observed that using the expanding ratio of 2 ($C = 512$, $D_{C} = 2C = 1024$) works better than the ratio value of 4 which is common in vision Transformers and MLPs. 
Increasing or reducing the number of GraphMLP layers $L$ hurts performance while using $L = 3$ performs best. 
The hidden size $C$ is important to determine the modeling ability. 
While increasing the $C$ from 128 to 512 (keeping the same MLP ratios), the MPJPE decreases from 50.2 $mm$ to 49.2 $mm$. 
Meanwhile, the number of parameters increases from 0.60M to 9.49M. 
The performance saturates when $C$ surpasses 512. 
Therefore, the optimal hyper-parameters for our model are $L = 3$, $C {= }512$, $D_{C} = 1024$, and $D_{S} = 256$, which are different from the original setting of MLP-Mixer~\cite{mlpmixer}. 
The reason for this may be the gap between vision and skeleton data, where the Human3.6M dataset is not diverse enough to train a large GraphMLP model. 

\begin{table}[t]
\centering
\scriptsize
\caption
{
Ablation study on various configurations of our approach. 
$L$ is the number of GraphMLP layers, $C$ is the hidden size, $D_{C}$ is the MLP dimension of CG-MLP, and $D_{S}$ is the MLP dimension of SG-MLP. 
}
\setlength{\tabcolsep}{3.33mm}
\begin{tabular}{cccccc}
\toprule [1pt]
$L$& $C$& $D_{C}$& $D_{S}$& Params (M) & MPJPE ($mm$) $\downarrow$ \\
\midrule
3& 512& 1024& 256& 9.49& \textbf{49.2} \\
\midrule [0.5pt]

3& 512& 1024& 128& 9.47& 49.7 \\
3& 512& 2048& 128& 12.62& 49.7 \\
3& 512& 2048& 256& 12.64& 49.6 \\

\midrule [0.5pt] 

2& 512& 1024& 256& 6.33& 50.2 \\
4& 512& 1024& 256& 12.65& 50.0 \\
5& 512& 1024& 256& 15.81& 50.3 \\
6& 512& 1024& 256& 18.97& 50.4 \\

\midrule [0.5pt]

3& 128& 256& 64& 0.60& 50.2 \\
3& 256& 512& 128& 2.38& 50.1 \\
3& 384& 768& 192& 5.35& 49.7 \\
3& 768& 1536& 384 & 21.31& 49.4 \\
\toprule [1pt]
\end{tabular}
\label{table:parameters}
\end{table}

\noindent \textbf{Input 2D Detectors.}
A high-performance 2D detector is vital in achieving accurate 3D pose estimation. 
Table~\ref{table:detectors} reports the performance of our model with ground truth 2D joints, and detected 2D joints from Stack Hourglass (SH)~\cite{newell2016stacked}, Detectron~\cite{videopose}, CPN~\cite{chen2018cascaded}, and HRNet~\cite{hrnet}. 
We can observe that our approach can produce more accurate results with a better 2D detector and is effective on different 2D estimators. 

\begin{table}[t]
  \centering
  \scriptsize
  \caption
  {
    Ablation study on different 2D detectors. 
  }
  \setlength{\tabcolsep}{5.73mm}
  \begin{tabular}{lcc}
  \toprule [1pt]
  Method &2D Mean Error &MPJPE ($mm$) $\downarrow$ \\
  \midrule [0.5pt]
  SH~\cite{newell2016stacked} &9.03 &56.9 \\ 
  Detectron~\cite{videopose} &7.77  &54.5 \\
  CPN~\cite{chen2018cascaded} &6.67 &49.2 \\

  2D Ground Truth &0 &\textbf{34.2} \\
  \toprule [1pt]
  \end{tabular}
  \label{table:detectors}
\end{table}

\begin{table}[t]
  \centering
  \scriptsize
  \caption
  {
    Ablation study on different network architectures. 
  }
  \setlength{\tabcolsep}{10.90mm}
  \begin{tabular}{lc}
  \toprule [1pt]
  Method &MPJPE ($mm$) $\downarrow$ \\
  \midrule [0.5pt]
  FCN~\cite{simplebaseline} &53.5 \\
  GCN~\cite{stgcn} &51.3 \\
  Transformer~\cite{transformer} &52.7 \\
  MLP-Mixer~\cite{mlpmixer} &52.0 \\
  Mesh Graphormer~\cite{meshgraphormer} &51.3 \\
  \midrule [0.5pt]
  Graph-Mixer& 50.4 \\
  Transformer-GCN& 51.5 \\
  GraphMLP &\textbf{49.2} \\

  \toprule [1pt]
  \end{tabular}
  \label{table:architecture}
\end{table}

\begin{figure*}[t]
  \centering
  \includegraphics[width=1.00\linewidth]{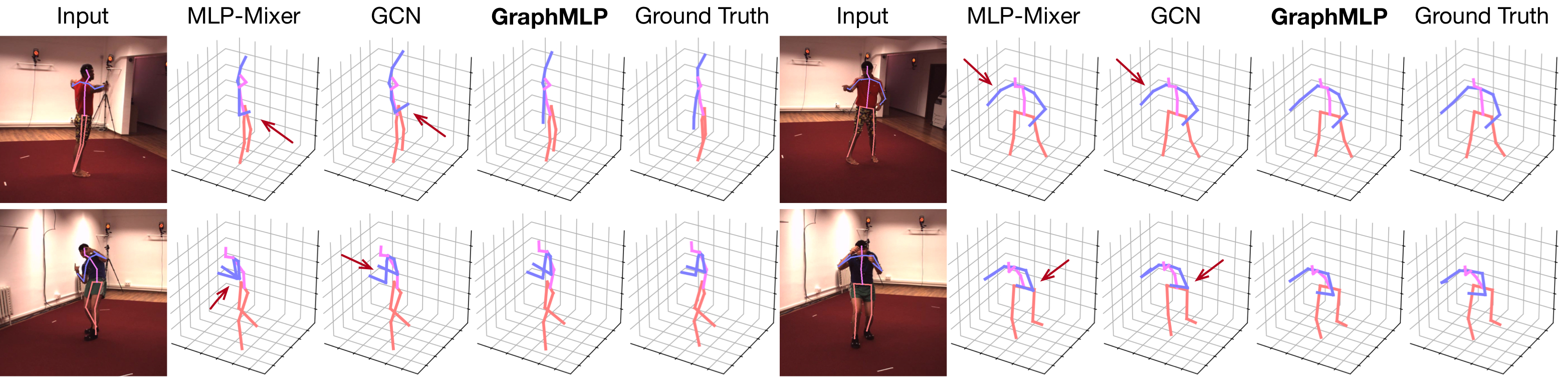}
  \caption
  {
  Qualitative comparison with MLP-Mixer~\cite{mlpmixer} and GCN~\cite{stgcn} for reconstructing 3D human poses on Human3.6M dataset. 
  Red arrows highlight wrong estimations. 
  }
  \label{fig:dataset}
\end{figure*}

\noindent \textbf{Network Architectures.}
To clearly demonstrate the advantage of the proposed GraphMLP, we compare our approach with various baseline architectures. 
The `Graph-Mixer' is constructed by replacing the linear layers in MLP-Mixer with GCN layers, and the `Transformer-GCN' is built by replacing the spatial MLPs in GraphMLP with multi-head self-attention blocks and adding a position embedding module. 
To ensure a consistent and fair evaluation, the parameters of these architectures (\emph{e.g.}, the number of layers, hidden size) remain the same. 
As shown in Table~\ref{table:architecture}, our proposed approach consistently surpasses all other architectures. 
For example, our approach can improve the GCN-based model~\cite{stgcn} from 51.3 $mm$ to 49.2 $mm$, resulting in relative 4.1\% improvements. 
It's worth noting that the MLP-Like model (`MLP-Mixer', `GraphMLP') outperforms the Transformer-based model (`Transformer', `Transformer-GCN'). 
This may be because a small number of tokens (\emph{e.g.}, 17 joints) are less effective for self-attention in learning long-range dependencies. 
Overall, these results confirm that GraphMLP can serve as a new and strong baseline for 3D human pose estimation. 
The implementation details of these network architectures can be found in the supplementary. 

\noindent \textbf{Network Design Options.}
Our approach allows the model to learn strong structural priors of human joints by injecting GCNs into the baseline model (\emph{i.e.}, MLP-Mixer~\cite{mlpmixer}). 
We also explore the different combinations of MLPs and GCNs to find an optimal architecture. 
As shown in Table~\ref{table:design}, the location matters of GCN are studied by five different designs: 
(i) The GCN is placed before spatial MLP. 
(ii) The GCN is placed after spatial MLP. 
(iii) The GCN is placed after channel MLP. 
(iv) The GCN and MLP are in parallel but using a spatial GCN block (process tokens in spatial dimension) in SG-MLP. 
(v) The GCN and MLP are in parallel. 
The results show that all these designs can help the model produce more accurate 3D poses, and using GCN and MLP in parallel achieves the best estimation accuracy. 
The illustrations of these design options can be found in the supplementary. 

\begin{table}[t]
\centering
\scriptsize
\caption
{
  Ablation study on different design options of combining MLPs and GCNs. 
  $^{*}$ means using spatial GCN block in SG-MLP. 
}
\setlength{\tabcolsep}{9.85mm}
\begin{tabular}{lc}
\toprule [1pt]
Method &MPJPE ($mm$) $\downarrow$ \\
\midrule [0.5pt]
Baseline &52.0 \\
\midrule [0.5pt]
GCN Before Spatial MLP &51.6 \\
GCN After Spatial MLP &50.6 \\
GCN After Channel MLP &50.9 \\
GCN and MLP in Parallel$^{*}$ &50.2 \\
GCN and MLP in Parallel &\textbf{49.2} \\
\toprule [1pt]
\end{tabular}
\label{table:design}
\end{table}

\begin{table}[t]
\centering
\scriptsize
\caption
{
  Ablation study on different components in GraphMLP. 
}
\setlength{\tabcolsep}{3.15mm}
\begin{tabular}{l|cc|c}
\toprule [1pt]
Method& SG-MLP& CG-MLP& MPJPE ($mm$) $\downarrow$ \\
\midrule [0.5pt]
Baseline &\xmark &
\xmark &52.0 \\
\midrule [0.5pt]
GraphMLP (SG-MLP) &\cmark &\xmark& 50.6 \\
GraphMLP (CG-MLP) &\xmark &\cmark &50.5 \\
GraphMLP &\cmark &\cmark &\textbf{49.2} \\

\toprule [1pt]
\end{tabular}
\label{table:components}
\end{table}

\noindent \textbf{Model Components.}
We also investigate the effectiveness of each component in our design. 
In Table~\ref{table:components}, the first row corresponds to the baseline model (\emph{i.e.}, MLP-Mixer~\cite{mlpmixer}) that does not use any GCNs in the model. 
The MPJPE is 52.0 $mm$. 
The rest of the rows show the results of replacing its spatial MLP with SG-MLP or channel MLP with CG-MLP by adding a GCN block into the baseline. 
It can be found that using our SG-MLP or CG-MLP can improve performance (50.6 $mm$ and 50.5 $mm$ respectively). 
When enabling both SG-MLP and CG-MLP, GraphMLP improves the performance over the baseline by a clear margin of 2.8 $mm$ (relatively 5.4\% improvements), indicating that the proposed components are mutually reinforced to produce more accurate 3D poses. 
These results validate the effectiveness of our motivation: combining modern MLPs and GCNs in a unified architecture for better 3D human pose estimation. 

\begin{figure}[t]
\centering
\includegraphics[width=1.0\linewidth]{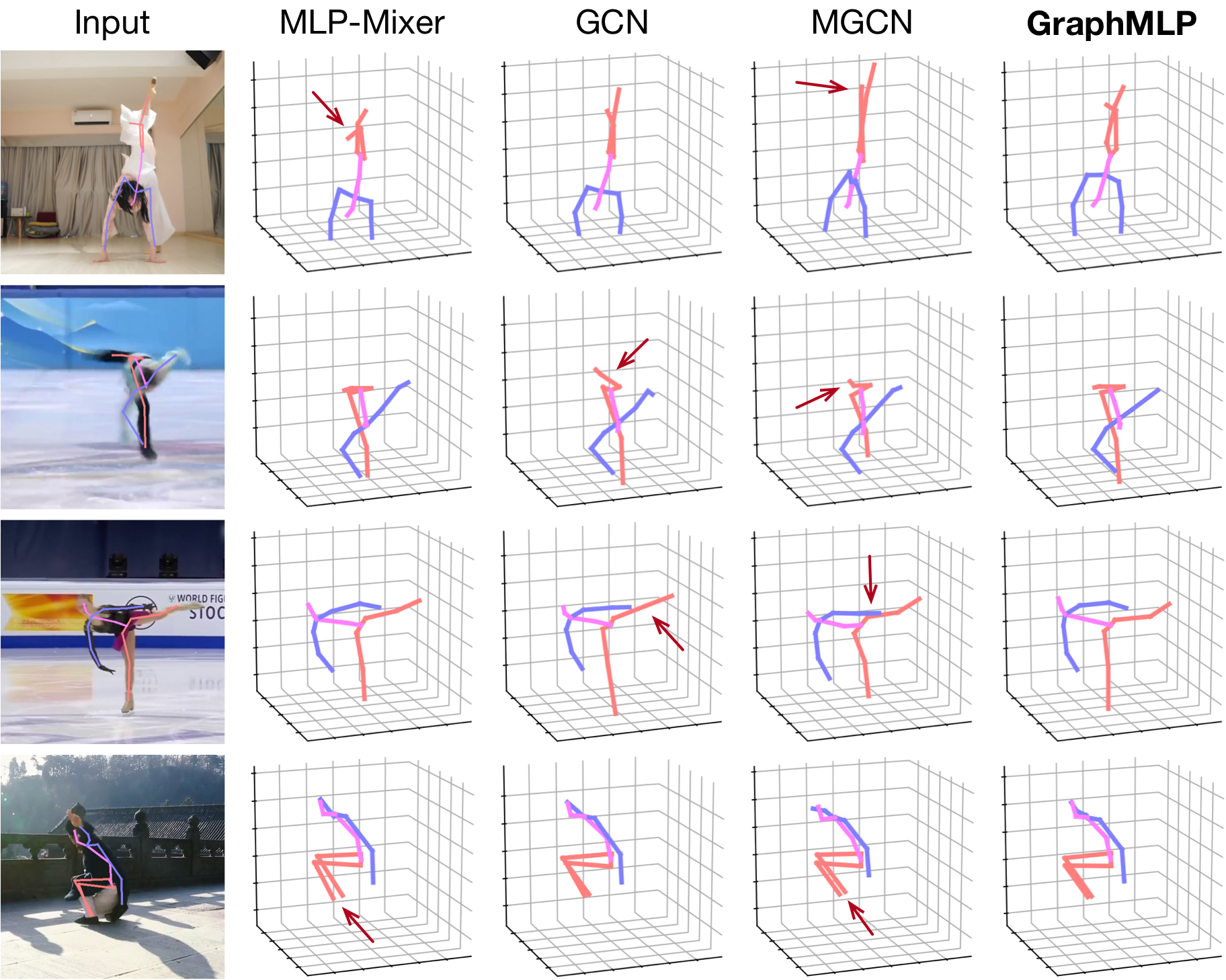}
\caption
{
  Qualitative comparison with MLP-Mixer~\cite{mlpmixer}, GCN~\cite{stgcn}, and MGCN~\cite{zou2021modulated} for reconstructing 3D human poses on challenging in-the-wild images. 
  Red arrows highlight wrong estimations. 
}
\label{fig:wild}
\end{figure}

\section{Qualitative Results}
Fig.~\ref{fig:dataset} presents the qualitative comparison among the proposed GraphMLP, MLP-Mixer~\cite{mlpmixer}, and GCN~\cite{stgcn} on the Human3.6M dataset. 
Furthermore, Fig.~\ref{fig:wild} presents the qualitative comparison among our GraphMLP and three other methods, namely MLP-Mixer~\cite{mlpmixer}, GCN~\cite{stgcn}, and MGCN~\cite{zou2021modulated}, on more challenging in-the-wild images.
Note that these actions from in-the-wild images are rare or absent from the training set of Human3.6M. 
GraphMLP, benefiting from its globality, locality, and connectivity, performs better and is able to predict accurate and plausible 3D poses. 
It indicates both the effectiveness and generalization ability of our approach. 
However, there are still some failure cases, where our approach fails to produce accurate 3D human poses due to large 2D detection error, half body, rare poses, and heavy occlusion, as shown in Fig.~\ref{fig:fail}. 
More qualitative results can be found in the supplementary. 

\section{Conclusion and Future Works}
In this paper, we propose a new graph-reinforced MLP-Like architecture, termed GraphMLP, which represents the first use of modern MLP dedicated to 3D human pose estimation. 
Our GraphMLP inherits the advantages of both MLPs and GCNs, making it a \textit{global-local-graphical} unified architecture without self-attention. 
Extensive experiments show that the proposed GraphMLP achieves state-of-the-art performance and can serve as a simple yet effective modern baseline for 3D human pose estimation in both single frame and video sequence. 
We also show that complex temporal dynamics can be effectively modeled in a simple way without explicitly temporal modeling, and the proposed GraphMLP in video reaches competitive results with fewer computational costs. 

Although the proposed approach has shown promising results, its performance is still limited by the quality of available datasets. 
High-quality 3D annotations rely on motion capture systems and are challenging and expensive to obtain. 
The most commonly used dataset for 3D human pose estimation, Human3.6M, lacks sufficient variations in terms of human poses, environments, and activities, which leads to models that do not generalize well to real-world scenarios. 
Domain adaptation and synthetic data are potential directions to improve the generalization ability of our approach. 
Moreover, since the MLPs and GCNs in our GraphMLP are straightforward, we look forward to incorporating more powerful MLPs or GCNs to further improve performance. 

\begin{figure}[t]
  \centering
  \includegraphics[width=1.0\linewidth]{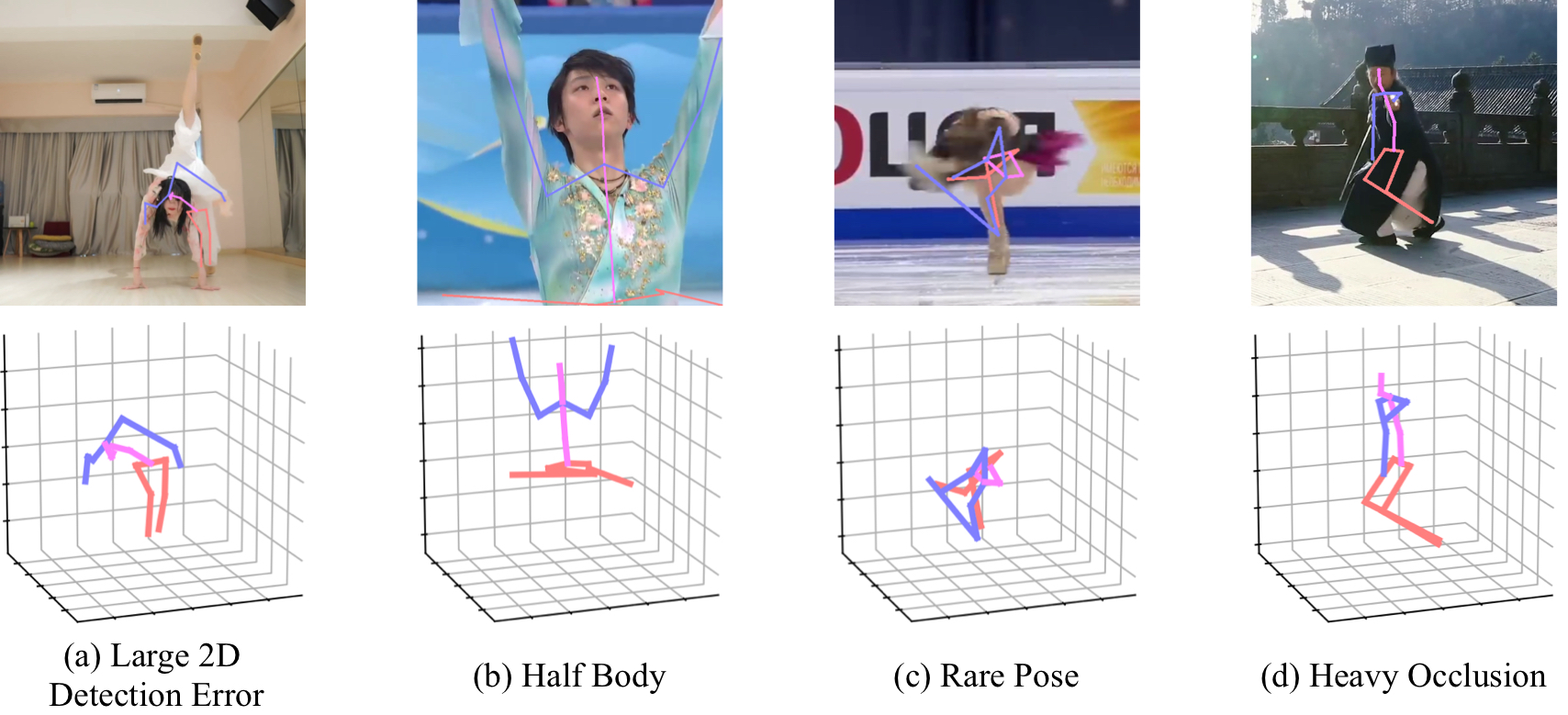}
  \caption
  {
    Challenging scenarios where GraphMLP fails to produce accurate 3D human poses.
  }
  \label{fig:fail}
\end{figure}
 
\section*{Acknowledgments}

This work was supported by the National Natural Science Foundation of China (No. 62373009), Natural Science Foundation of Guangdong Province (No. 2024A1515012089), Natural Science Foundation of Shenzhen (No. JCYJ20230807120801002), Shenzhen Innovation in Science and Technology Foundation for The Excellent Youth Scholars (No. RCYX20231211090248064), the Fundamental Research Funds for the Central Universities, Peking University, the MUR PNRR project FAIR (PE00000013) funded by the NextGenerationEU, and the EU Horizon project ELIAS (No. 101120237). 

{\small
\bibliographystyle{ieeenat_fullname}
\bibliography{ref}
}

\clearpage
\appendix
{\noindent\Large\textbf{Supplementary Material}}
\newline

This supplementary material contains the following details:
(1) Detailed description of multi-layer perceptrons (see Sec.~\ref{sec:mlp}). 
(2) Additional implementation details (see Sec.~\ref{sec:details}). 
(3) Additional ablation studies (see Sec.~\ref{sec:ablation}).
(4) Additional qualitative results (see Sec.~\ref{sec:qualitative}). 

\section{Multi-Layer Perceptrons}
\label{sec:mlp}
In Sec.~\ref{sec:preliminary} of our main manuscript, we give a brief description of the MLP-Mixer layer~\cite{mlpmixer} which is defined as below:
\begin{align}
  X^{\prime}_{\ell} &= X_{\ell-1} + \operatorname{Spatial-MLP}(\operatorname{LN}(X_{\ell-1}^{\top}))^{\top}, 
  \label{equ:mlpmixer_spatial_supp} \\
  X_{\ell} &= X^{\prime}_{\ell} + \operatorname{Channel-MLP}(\operatorname{LN}(X^{\prime}_{\ell})).
  \label{equ:mlpmixer_channel_supp}
\end{align}

If considering more details about the spatial and channel MLPs, Eq.~\eqref{equ:mlpmixer_spatial_supp} and Eq.~\eqref{equ:mlpmixer_channel_supp} can be further defined as:
\begin{align}
  X^{\prime}_{\ell} &= X_{\ell-1} + \left(W_{2} \sigma\left(W_{1}(\operatorname{LN}(X_{\ell-1}^{\top}))\right)\right)^{\top}, \label{equ:mlpmixer_spatial_supp_1}\\
  X_{\ell} &= X^{\prime}_{\ell} + W_{4} \sigma\left(W_{3}(\operatorname{LN}(X_{\ell-1}))\right)),
\end{align}
where $\sigma$ is the GELU activation function~\cite{hendrycks2016gaussian}.
$W_{1} \in \mathbb{R}^{N \times D_{S}}$ and $W_{2} \in \mathbb{R}^{D_{S} \times N}$ are the weights of two linear layers in the $\operatorname{Spatial-MLP}(\cdot)$. 
$W_{3} \in \mathbb{R}^{C \times D_{C}}$ and $W_{2} \in \mathbb{R}^{D_{C} \times C}$ are the weights of two linear layers in the $\operatorname{Channel-MLP}(\cdot)$. 

For GraphMLP in video, LN is additionally applied after every fully-connected layer in $\operatorname{Spatial-MLP}(\cdot)$, Eq.~\eqref{equ:mlpmixer_spatial_supp_1} is modified to:
\begin{equation}
  X^{\prime}_{\ell} = X_{\ell-1} + \left(\operatorname{LN}\left(W_{2} \sigma\left(\operatorname{LN}\left(W_{1}(\operatorname{LN}(X_{\ell-1}^{\top}))\right)\right)\right)\right)^{\top}. 
\end{equation}

\section{Additional Implementation Details}
\label{sec:details}
In Sec.~\ref{sec:ablation} of our main manuscript, we conduct extensive ablation studies on different network architectures and model variants of our GraphMLP. 
Here, we provide implementation details of these models. 

\begin{figure*}[b]
\centering
\includegraphics[width=1.0\linewidth]{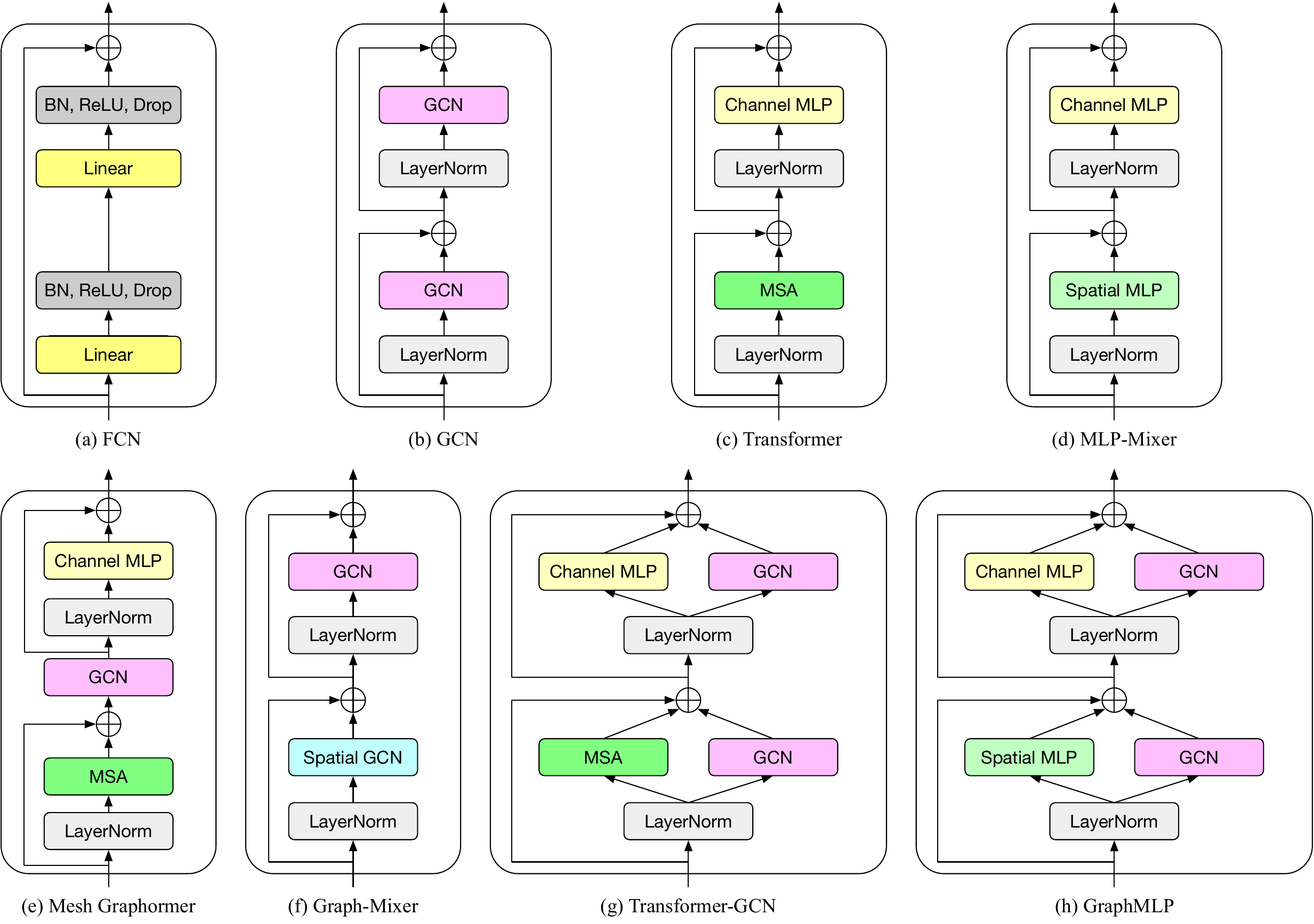}
\caption
{ 
   Different network architectures. 
}
\label{fig:supp_network}
\end{figure*}

\noindent \textbf{Network Architectures.}
In Table~\ref{table:architecture} of our main paper, we report the results of various network architectures. 
Here, we provide illustrations in Fig.~\ref{fig:supp_network} and implementation details as follows:
\begin{itemize}
  \item FCN: FCN~\cite{simplebaseline} is a conventional MLP whose building block contains a linear layer, followed by batch normalization, dropout, and a ReLU activation. 
  We follow their original implementation and use their code~\cite{code_baeline} to report the performance. 
  \item GCN: We remove the spatial MLP and the channel MLP in our SG-MLP and CG-MLP, respectively. 
  \item Transformer: We build it by using a standard Transformer encoder which is the same as ViT~\cite{vit}. 
  \item MLP-Mixer: It is our baseline model that has the same architecture as~\cite{mlpmixer}. We build it by replacing multi-head attention blocks with spatial MLPs and removing the position embedding module in the Transformer. 
  \item Mesh Graphormer: Mesh Graphormer~\cite{meshgraphormer} is the most relevant study to our approach that focuses on combining self-attentions and GCNs in a Transformer model for mesh reconstruction. 
  Instead, our GraphMLP focuses on combining modern MLPs and GCNs to construct a stronger architecture for 3D human pose estimation. 
  We follow their design to construct a model by adding a GCN block after the multi-head attention in the Transformer. 
  \item Graph-Mixer: We replace linear layers in MLP-Mixer with GCN layers. 
  \item Transformer-GCN: We replace spatial MLPs in GraphMLP with multi-head self-attention blocks and add a position embedding module before Transformer-GCN layers. 
  \item GraphMLP: It is our proposed approach. Please refer to Fig.~\ref{fig:overview} of our main paper. 
\end{itemize}

\begin{figure*}[t]
\centering
\includegraphics[width=0.59\linewidth]{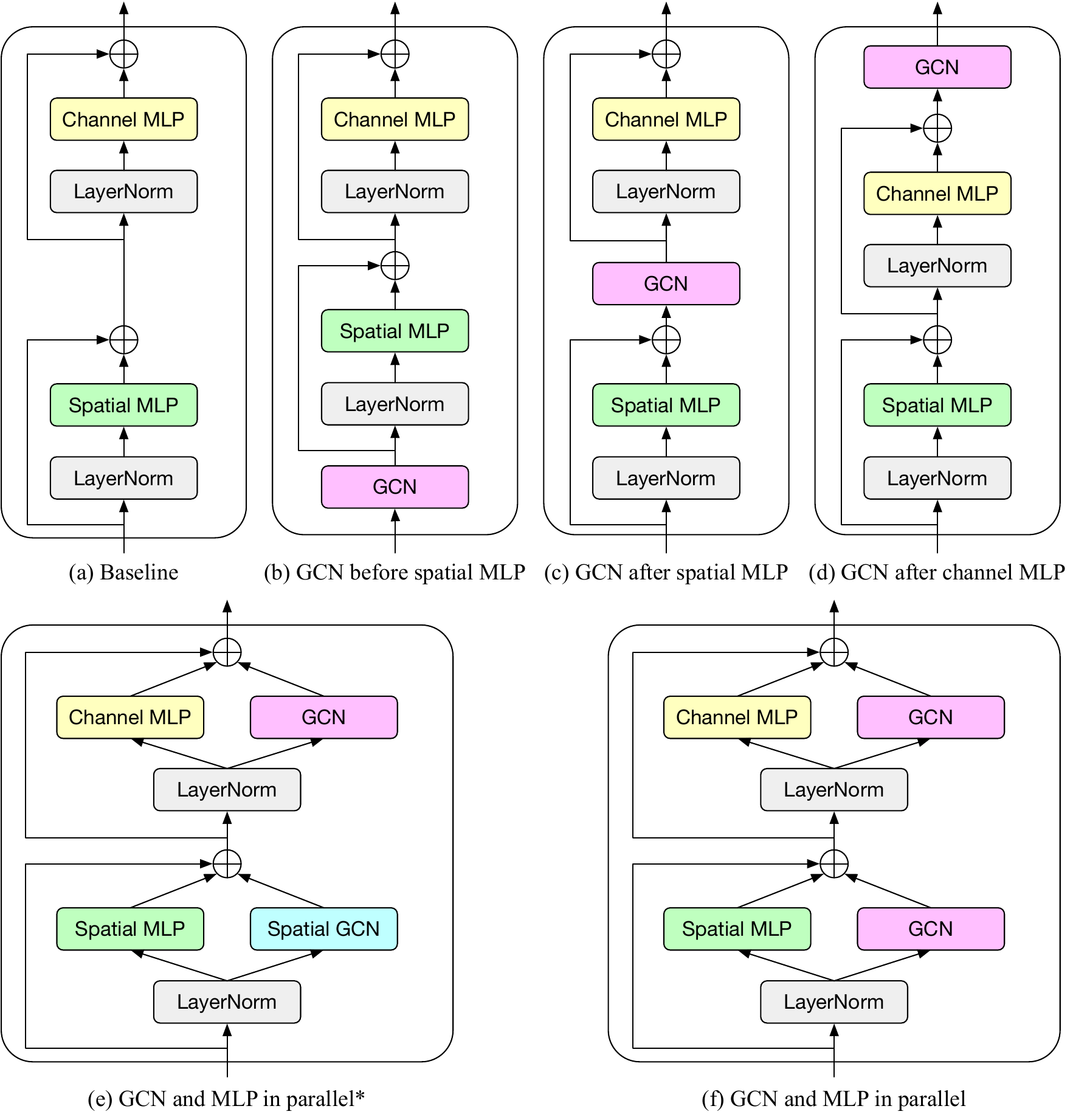}
\caption
{
   Different design options of the GraphMLP layer. 
}
\label{fig:supp_design}
\end{figure*}

\noindent \textbf{Network Design Options.}
In Fig.~\ref{fig:supp_design}, we graphically illustrate five different design options of the GraphMLP layer as mentioned in Table~\ref{table:design} of our main paper. 
Fig.~\ref{fig:supp_design} (e) shows that we adopt the design of GCN and MLP in parallel but use a spatial GCN block in SG-MLP. 
The spatial GCN block processes tokens in the spatial dimension, which can be calculated as:
\begin{equation}
  \begin{aligned}
  X^{\prime}_{\ell} = X_{\ell-1} &+ \operatorname{Spatial-MLP}(\operatorname{LN}(X_{\ell-1}^{\top}))^{\top} \\
  & + \operatorname{Spatial-GCN}(\operatorname{LN}({X_{\ell-1}^{\top}}))^{\top}.
\end{aligned}
\end{equation}

\begin{table}[t]
\centering
\scriptsize
\caption
{Ablation study on different design options of transposition. 
}
\setlength{\tabcolsep}{7.15mm}
\begin{tabular}{lc}
\toprule
Method &MPJPE ($mm$) $\downarrow$ \\
\midrule
MLP-Mixer, Transposition After LN &52.6 \\
MLP-Mixer, Transposition Before LN &\textbf{52.0} \\

\midrule
GraphMLP, Transposition After LN &49.7 \\
GraphMLP, Transposition Before LN &\textbf{49.2} \\

\toprule
\end{tabular}
\label{table:transposition}
\end{table}

\begin{figure*}[t]
\centering
\includegraphics[width=0.59\linewidth]{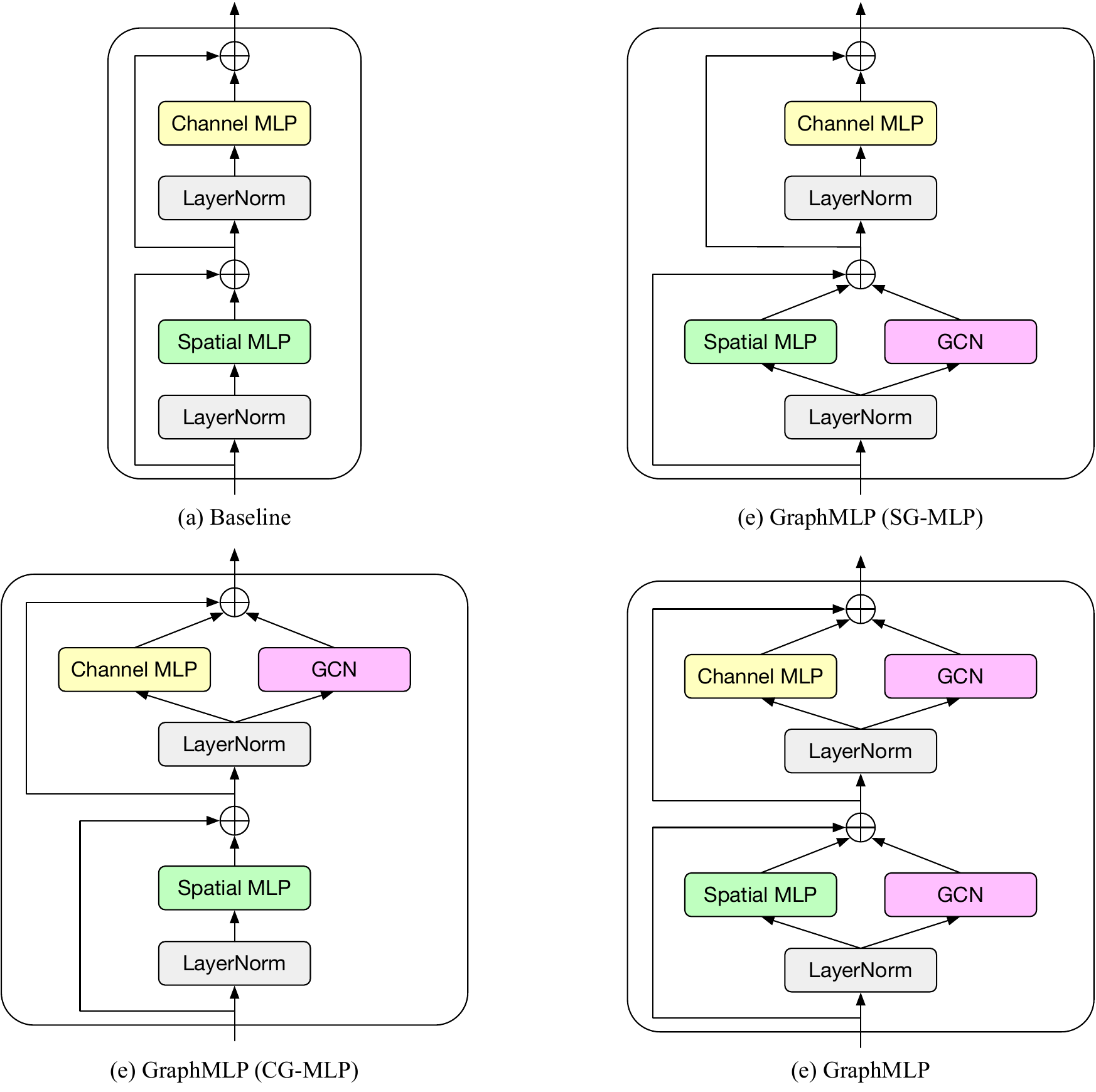}
\caption
{
   The model components we have studied for building our proposed GraphMLP. 
}
\label{fig:supp_component}
\end{figure*}

\noindent \textbf{Model Components.}
In Table~\ref{table:components} of our main paper, we investigate the effectiveness of each component in our design. 
Here, we provide the illustrations of these model variants in Fig.~\ref{fig:supp_component}. 

\section{Additional Ablation Studies}
\label{sec:ablation}
\noindent \textbf{Transposition Design Options.}
As mentioned in Sec.~\ref{sec:preliminary} of our main manuscript, we transpose the tokens before LN in the spatial MLP, and therefore the LN normalizes tokens along the spatial dimension. 
Here, we investigate the influence of transposition design options in Table~\ref{table:transposition}. 

The `Transposition Before LN' can be formulated as:
\begin{equation}
  \begin{aligned}
  X^{\prime}_{\ell} = X_{\ell-1} &+ \operatorname{Spatial-MLP}(\operatorname{LN}(X_{\ell-1}^{\top}))^{\top} \\
  & + \operatorname{GCN}(\operatorname{LN}({X_{\ell-1}^{\top}})^{\top}). 
\end{aligned}
\end{equation}

The `Transposition After LN' can be written as:
\begin{equation}
  \begin{aligned}
  X^{\prime}_{\ell} = X_{\ell-1} &+ \operatorname{Spatial-MLP}(\operatorname{LN}(X_{\ell-1})^{\top})^{\top} \\
  & + \operatorname{GCN}(\operatorname{LN}({X_{\ell-1}})). 
\end{aligned}
\end{equation}

From Table~\ref{table:transposition}, the results show that performing transposition before LN brings more benefits in both MLP-Mixer and our GraphMLP models. 
Note that it is different from the original implementation of MLP-Mixer, which uses transposition after LN. 

\section{Additional Qualitative Results}
\label{sec:qualitative}
\begin{figure*}[htb]
\centering
\includegraphics[width=0.90\linewidth]{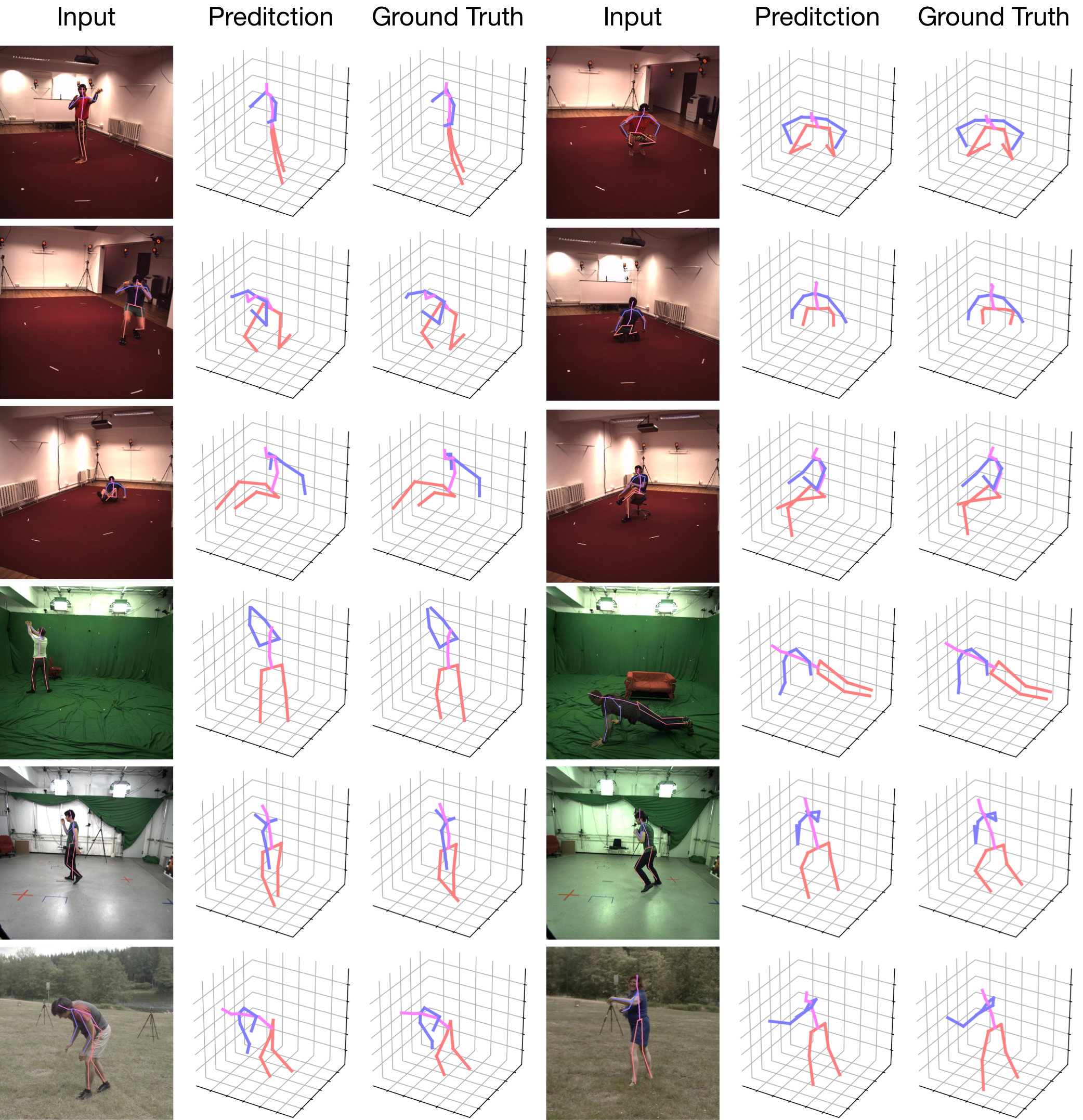}
\caption
{Visualization results of our approach for reconstructing 3D human poses on Human3.6M dataset (top three rows) and MPI-INF-3DHP dataset (bottom three rows). 
}
\label{fig:supp_dataset}
\end{figure*}

\begin{figure*}[t]
\centering
\includegraphics[width=0.90\linewidth]{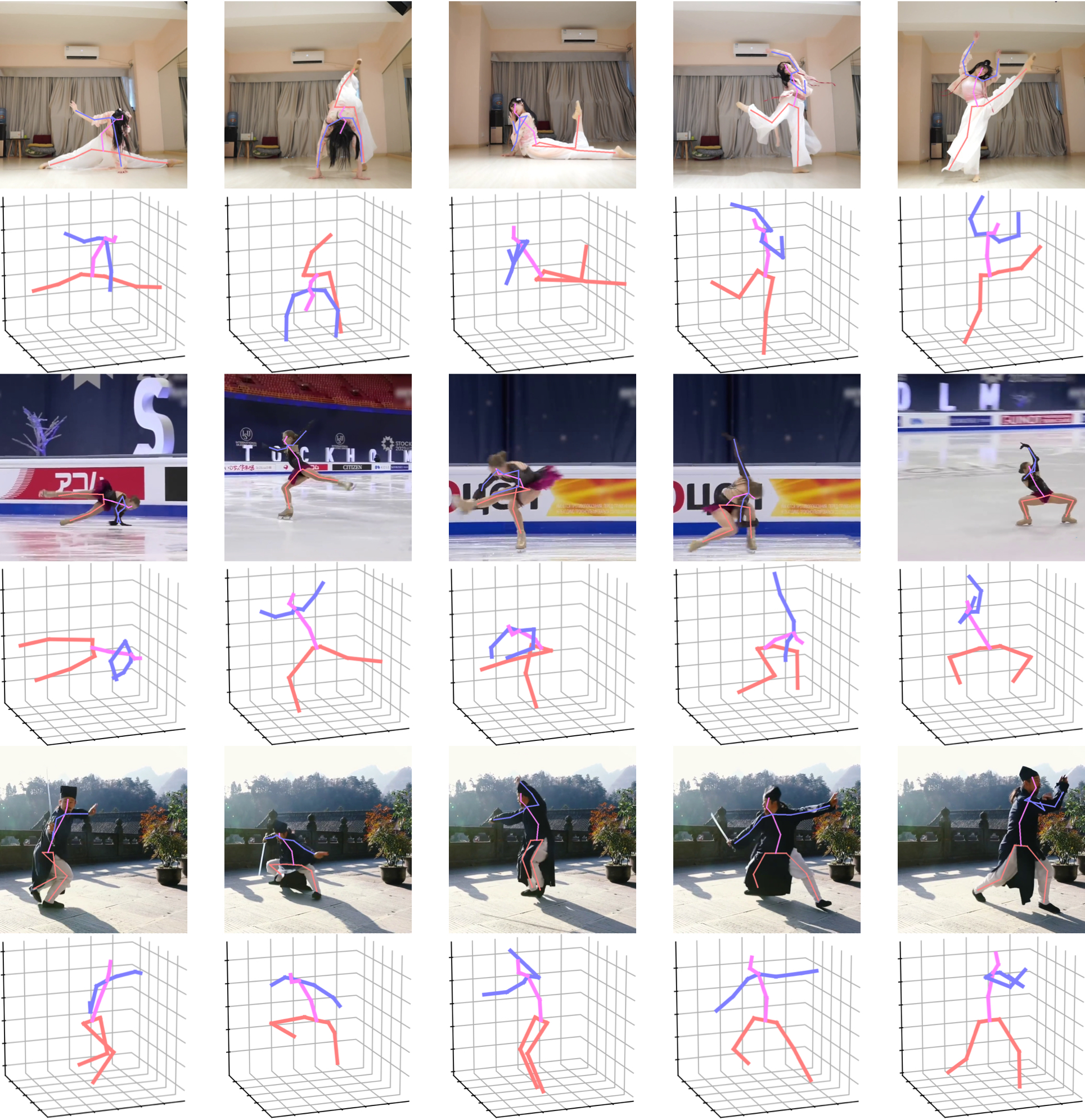}
\caption
{Visualization results of the proposed GraphMLP for reconstructing 3D human poses on challenging in-the-wild images. 
}
\label{fig:supp_wild}
\end{figure*}

Fig.~\ref{fig:supp_dataset} shows qualitative results of the proposed GraphMLP on Human3.6M and MPI-INF-3DHP datasets. 
The Human3.6M is an indoor dataset (top three rows), and the test set of MPI-INF-3DHP contains three different scenes: studio with green screen (GS, fourth row), studio without green screen (noGS, fifth row), and outdoor scene (Outdoor, sixth row). 
Moreover, Fig.~\ref{fig:supp_wild} shows qualitative results on challenging in-the-wild images. 
We can observe that our approach is able to predict reliable and plausible 3D poses in these challenging cases. 

\end{document}